\documentclass[preprint,12pt,authoryear]{elsarticle}
\let\today\relax
\makeatletter
\def\ps@pprintTitle{%
    \let\@oddhead\@empty
    \let\@evenhead\@empty
    \def\@oddfoot{\footnotesize\itshape
         {Submitted preprint} \hfill\today}%
    \let\@evenfoot\@oddfoot
    }
\makeatother

\usepackage{amssymb}
\usepackage{amsmath}

\usepackage{graphicx}
\usepackage{subcaption}
\usepackage[export]{adjustbox}
\usepackage[a4paper, left=2cm, right=2cm, top=1cm, bottom=2cm]{geometry}
\usepackage{amsmath}
\usepackage{amssymb}
\usepackage{nomencl}
\usepackage{multicol}
\usepackage{framed}
\usepackage{float}    
\usepackage{afterpage}     
\usepackage{bm}
\usepackage{multirow}
\usepackage{xcolor}
\usepackage{tcolorbox}
\usepackage{hyperref}
\usepackage{cleveref}
\usepackage{multirow}
\usepackage{booktabs}
\usepackage{tabularx}
\usepackage{siunitx}
\usepackage{tcolorbox}
\usepackage{algorithm}
\usepackage{algorithmicx}
\usepackage{algpseudocode}
\usepackage[T1]{fontenc}
\usepackage{textcomp}
\usepackage{lmodern}
\newtcolorbox[auto counter, number within=section]{examplebox}[2][]{colback=white,
colframe=purple!60!black, fontupper=\ttfamily\small, title=Example~\thetcbcounter: #2,#1}
\tcbuselibrary{listings, breakable}
\usepackage{listings}
\crefname{equation}{Eq.}{Equations}
\Crefname{equation}{Eq.}{Equations}
\begin{document}

\begin{frontmatter}

\title{Autonomous Control Leveraging LLMs: An Agentic Framework for Next-Generation Industrial Automation} 

\author[Imperial]{Javal Vyas}
\author[Imperial]{Mehmet Mercangöz}
\affiliation[Imperial]{organization={Autonomous Industrial Systems Lab,Imperial College London},
            addressline={Imperial College Rd, South Kensington Campus},
            city={London},
            postcode={SW7 2AZ},
            country={United Kingdom}}


\begin{abstract}
The increasing complexity of modern chemical processes, coupled with workforce shortages and intricate fault scenarios, demands novel automation paradigms that blend symbolic reasoning with adaptive control. In this work, we introduce a unified agentic framework that leverages large language models (LLMs) for both discrete fault-recovery planning and continuous process control within a single architecture. We adopt Finite State Machines (FSMs) as interpretable operating envelopes: an LLM-driven planning agent proposes recovery sequences through the FSM, a Simulation Agent executes and checks each transition, and a Validator–Reprompting loop iteratively refines invalid plans. In Case Study 1, across 180 randomly generated FSMs of varying sizes (4–25 states, 4–300 transitions), GPT-4o and GPT-4o-mini achieve 100 \% valid-path success within five reprompts—outperforming open-source LLMs in both accuracy and latency. In Case Study 2, the same framework modulates dual-heater inputs on a laboratory TCLab platform (and its digital twin) to maintain a target average temperature under persistent asymmetric disturbances. Compared to classical PID control, our LLM-based controller attains similar performance, while ablation of the reprompting loop reveals its critical role in handling nonlinear dynamics. We analyze key failure modes—such as instruction-following lapses and coarse ODE approximations. Our results demonstrate that, with structured feedback and modular agents, LLMs can unify high-level symbolic planning and low-level continuous control, paving the way toward resilient, language-driven automation in chemical engineering.
\end{abstract}
\begin{keyword}
Autonomous Control\sep Large Language Models (LLMs) \sep Agentic AI Framework \sep Industrial Automation \sep Zero-Shot Learning


\end{keyword}

\end{frontmatter}



\section{Introduction}

Industrial automation is approaching a critical inflection point\citep{Pantelides2024}. For decades, operational systems have relied on rule-based logic, deterministic models, and tightly structured control architectures. While these approaches have proven robust in well-defined settings, they increasingly fall short under the rising complexity of modern industrial operations. Workforce shortages, ageing infrastructure, distributed asset networks, and growing variability in inputs and market conditions are exposing the limitations of conventional automation. At the same time, the convergence of industrial digitization and recent breakthroughs in artificial intelligence are catalyzing a shift toward more adaptive, data-driven, and interpretable control frameworks \citep{borghesan2022unmanned}.

A central enabler of this transformation is the emergence of \textbf{Large Language Models (LLMs)}. These models exhibit remarkable capabilities in reasoning, adaptation, and natural language understanding—making them fundamentally different from traditional automation tools. LLMs can generalize across tasks, construct plans in loosely defined problem spaces, and interact with humans in expressive and intuitive ways. As such, they are uniquely suited for applications where symbolic reasoning, fault diagnosis, and flexible control must coexist under uncertainty and partial observability.

In the context of industrial autonomy, LLMs open new opportunities to emulate high-level cognitive functions traditionally reserved for human operators and engineers. This includes the ability to interpret unstructured information, propose corrective actions in response to unanticipated events, and coordinate multi-step procedures that extend beyond pre-programmed responses. However, leveraging LLMs in safety-critical environments also presents several open challenges. Key questions include: \textit{Can LLMs operate robustly under fault conditions? Can they synthesize valid control strategies or recovery plans in dynamic, partially observable settings? How can symbolic planning capabilities be grounded in the physics and constraints of real-world plant dynamics?}

Addressing these questions will require new architectures that tightly couple LLM-based reasoning with mechanistic models, domain knowledge, and low-level control systems. The integration of LLMs into industrial workflows is not merely an interface enhancement—it is a foundational shift that could redefine how autonomy is conceived, developed, and validated in complex engineered systems.

\subsection{Motivation for autonomous industrial systems}

The urgency for autonomy is growing due to several converging factors. First, demographic shifts threaten the availability of skilled operators: a 2022 U.S. NAM report estimates that over 2.1 million manufacturing jobs could go unfilled by 2030 due to retirements and labor shortages \citep{nam2021pressrelease, deloitte2021manufacturing}. Additionally, operational errors are often rooted not in hardware faults but in \textbf{procedural breakdowns}. Operators frequently rely on memory or informal heuristics under stress, leading to deviation from official procedures and raising the risk of human error.

Particularly concerning is the challenge of handling \textbf{rare fault conditions}. Due to their infrequency, operators lack experience and must act under time pressure, often amidst alarm floods that hinder clear judgment \citep{alarmfloods, alarmflooding2}. Despite the proliferation of sensors and digital infrastructure through Industry 4.0 initiatives, many systems lack adaptive intelligence to translate this data into effective real-time decisions.

We argue that LLMs, when embedded in modular agentic frameworks, are well-positioned to address these gaps by offering both \textbf{flexible assistance} and \textbf{full autonomy}, depending on operational needs. They can monitor plant states, interpret symbolic abstractions (like FSMs), generate feasible plans, and validate their actions in-situ.

\subsection{Why LLMs? Reasoning, adaptation, and control}

LLMs represent a new class of intelligent systems with two key advantages over classical automation paradigms: \textbf{context-aware reasoning with self-consistency}, and \textbf{expressive interfacing across modalities}.

Unlike rule-based or supervised models, LLMs dynamically reason during inference and can revise their decisions mid-generation if contradictions are detected \citep{COT, selfConsistency}. This makes them inherently more robust in environments where fault conditions are rare, ambiguous, or not explicitly modeled during training. Techniques like chain-of-thought (CoT) prompting and self-consistency sampling further enhance their ability to perform multi-step symbolic reasoning \citep{COT, RLHF}.

Furthermore, LLMs can process free-form inputs—text logs, sensor descriptions, natural language queries—and produce semantically rich outputs without rigid formatting constraints \citep{Gill2025}. This bridges the gap between human operators and autonomous systems, enabling seamless integration into plant environments where data exists in diverse formats.


\subsection{Problem scope and contributions}

The integration of Large Language Models (LLMs) into industrial control presents new opportunities for building interpretable, adaptive, and fault-resilient automation systems. However, their application in real-time plant operation remains underexplored, particularly in hybrid settings that require both symbolic reasoning and grounded physical control.

This paper presents a preliminary investigation into the potential of Large Language Models (LLMs) to operate effectively within an agentic architecture for autonomous industrial control. We focus on two complementary dimensions of autonomy: (i) symbolic planning over structured abstractions, and (ii) continuous control in dynamic, real-time environments. To isolate and assess each capability, we design two targeted case studies—each probing a distinct facet of the LLM's decision-making performance. Through these studies, we aim to evaluate whether LLM-based agents can reason, adapt, and recover from unanticipated events without human intervention, and to identify the architectural elements necessary for grounding high-level symbolic reasoning in low-level process dynamics.

\begin{itemize}
    \item \textbf{Case study I: Symbolic recovery planning via FSM traversal.} We formalize fault handling as a path-planning problem over a Finite State Machine (FSM) and evaluate the ability of LLM agents to generate valid recovery trajectories from a fault state to a normal operation state. A multi-agent loop involving validation and reprompting is used to correct invalid transitions and improve planning robustness across FSMs of varying complexity.
    
    \item \textbf{Case study II: Continuous control under disturbances.} Using a dual-heater temperature control setup (TCLab), we test whether LLMs can generate real-time power inputs to maintain a temperature setpoint in the presence of asymmetrical faults. We compare LLM-based control against classical PID baselines under varying model configurations and deployment environments.
\end{itemize}

\noindent
The primary contributions of this work are as follows:
\begin{enumerate}
    \item We demonstrate that LLM-based agents, when integrated into an agentic framework incorporating validation and reprompting components, are capable of performing both symbolic reasoning and continuous control tasks. In structured benchmark settings, their performance approaches that of classical methods, highlighting the potential of LLMs to serve as adaptive, interpretable components within autonomous control systems.
    
    \item We identify key limitations in current LLM behavior, particularly with respect to instruction adherence and reasoning over physical system models. These limitations are quantified through systematic error analysis and control performance metrics, providing insight into failure modes that must be addressed for reliable deployment in safety-critical applications.

\end{enumerate}

The remainder of this paper is organized as follows. Section~\ref{sec:related_work} reviews related work on LLMs in industrial control, planning capabilities of LLMs, FSM based modeling, use of digital twins and prompting strategies. Section~\ref{sec:methodology} introduces the proposed multi-agent framework, detailing its architecture, validation loop, and design principles. Section~\ref{sec:case_studies} presents two case studies that evaluate the framework in complementary domains: symbolic planning over FSMs and continuous temperature control under disturbances. For each, we describe the experimental setup, agent configurations, and performance metrics. Section~\ref{sec:discussion} synthesizes insights across the two studies, highlighting strengths and current limitations of LLM-based autonomy. Finally, Section~\ref{sec:conclusion} concludes the paper with a discussion of future research directions, including integration with digital twins and applications to real-world industrial fault management.

\section{Related work}
\label{sec:related_work}
This section reviews prior work across five interrelated areas: LLM-based industrial control, planning capabilities of LLMs, FSM-based modeling and diagnosis, integration of digital twins and prompting strategies. Together, these domains frame the motivation and novelty of our bifurcated agentic framework for symbolic planning and physical control using LLMs.

\subsection{LLMs in industrial control}
The use of large language models in industrial control is an emerging area, with promising demonstrations across HVAC regulation, PLC programming, and flexible manufacturing. Song \emph{et al.}~\citep{song2023llmplanner} formulated HVAC control as a language modeling task and showed that GPT-4, when prompted with few-shot examples, could match reinforcement learning baselines in controlling building temperature. Fakih \emph{et al.}~\citep{Fakih2024} introduced LLM4PLC, a pipeline for generating IEC 61131-3 compliant PLC programs from natural language, using verifier feedback and grammar-aware prompting to improve correctness. Similarly, Xia \emph{et al.}~\citep{Xia2023, Xia2024} proposed LLM-driven automation pipelines that interpret digital twin state and synthesize production plans, showcasing end-to-end control execution from textual goals. These efforts establish the feasibility of LLMs in structured control settings, but often focus on static mappings (text-to-code) or simple policy generation. Our work differs by deploying LLMs in a closed-loop inference–validation–execution setting, and measuring adaptive performance under continuous disturbances and symbolic reasoning constraints.

\subsection{LLM-based planning and graph traversal}
LLMs exhibit emerging planning abilities, yet their reliability diminishes with increasing plan horizon and graph complexity. Valmeekam \emph{et al.}~\citep{valmeekam2022large} highlight that even with structured prompts, LLMs hallucinate transitions in symbolic planning tasks. In response, hybrid planners such as LLM-Planner~\citep{song2023llmplanner} and neural-symbolic frameworks~\citep{hou2025neural} propose grounding LLM outputs in formal domain models or modularizing planning and execution. Related works explore LLM performance in graph-structured domains~\citep{jin_large_2024, agrawal_can_2024}, with studies showing that traversal accuracy is sensitive to graph encoding and prompt structure. Our work contributes to this literature by integrating a planning agent with an explicit correction mechanism over FSM graphs. We show that structured reprompting effectively reduces path invalidity and improves planning accuracy across FSM complexities. This demonstrates that LLMs, when paired with validation layers, can support long-horizon symbolic reasoning.

\subsection{Finite state machines for fault modeling and recovery}
Finite State Machines (FSMs) have long served as formal models for discrete-event control and fault recovery in cyber-physical systems. Classical supervisory control theory, as reviewed in~\citep{fsm_for_plc}, employs FSMs for enforcing safety and recovery logic in PLC-based controllers. Recent works extend FSMs to include dynamic observers, such as Zhang \emph{et al.}~\citep{Zhang2021}, who use extended FSMs with super-twisting observers for real-time sensor fault detection in motor drives. Chen \emph{et al.}~\citep{Chen2024} apply integer linear programming to construct diagnoser automata that guarantee fault identifiability given partial observability constraints. These methods offer structured, verifiable fault detection but often lack adaptability in unanticipated scenarios. We adopt FSMs not merely as control logic, but as symbolic abstractions of operating envelopes, enabling LLMs to reason about recovery paths while retaining interpretability. Unlike prior works that assume hand-coded transitions, our agents dynamically generate valid state sequences under FSM-defined constraints, augmented by corrective validation.

\subsection{Digital twins and LLM integration}
Digital twins provide simulation environments for validating control policies under varying conditions. Recent reviews~\citep{Yang2025} highlight the potential of LLMs to enhance digital twin workflows via description (data annotation), prediction (context-aware modeling), and prescription (control synthesis). Sheng \emph{et al.}~\citep{Sheng2024} fuse digital twin data with LLMs in federated learning setups, illustrating synergy in distributed environments. However, most efforts remain at the level of descriptive analytics or offline planning. We integrate digital twins as online simulators in a closed-loop architecture with LLM agents, enabling real-time validation of control trajectories. This bridges the gap between simulated insight and actionable control, and demonstrates that LLMs can meaningfully operate in time-synchronized decision loops when paired with twin environments.

\subsection{Prompting and iterative correction strategies}
Recent advances in LLM prompting have extended beyond static in-context learning to dynamic, feedback-based reasoning. Chain-of-Thought (CoT) prompting~\citep{COT} and its extensions such as Tree-of-Thought (ToT)~\citep{TreeOfThought} and Graph-of-Thought (GoT)~\citep{GraphOfThought} show that guiding LLMs to reason through intermediate steps improves accuracy in planning and decision tasks. Complementing these, frameworks like ReAct~\citep{React} and Reflexion~\citep{REFLEXION} introduce iterative correction loops using self-generated feedback, while reprompting-based pipelines enable adaptation to constraint violations and hallucinations~\citep{reprompting}. Our framework builds on these ideas by instantiating a validator–reprompting agent loop for both symbolic plans and control actions. We extend the reprompting approach to include multiple constraint layers (e.g., temperature validity, power bounds) and demonstrate its stabilizing effect in both discrete and continuous domains.

While prior studies have addressed either symbolic planning or control synthesis in isolation, our work advances a unified agentic architecture that modularly separates these responsibilities into dedicated planning and control agents, each augmented by domain-specific validation and reprompting mechanisms. Although our current case studies focus on these components independently, the overarching goal is to deploy FSMs as interpretable operating envelopes that mediate transitions between control regimes. By embedding LLMs within this structured framework, we demonstrate improved reliability in both symbolic planning and closed-loop control. This approach lays the groundwork for autonomous systems that integrate interpretability, flexibility, and self-correction—capabilities not jointly realized in existing LLM-based frameworks for industrial automation.
\section{Methodology}
\label{sec:methodology}

\begin{figure}[ht]
    \centering
    \includegraphics[width=0.9\textwidth]{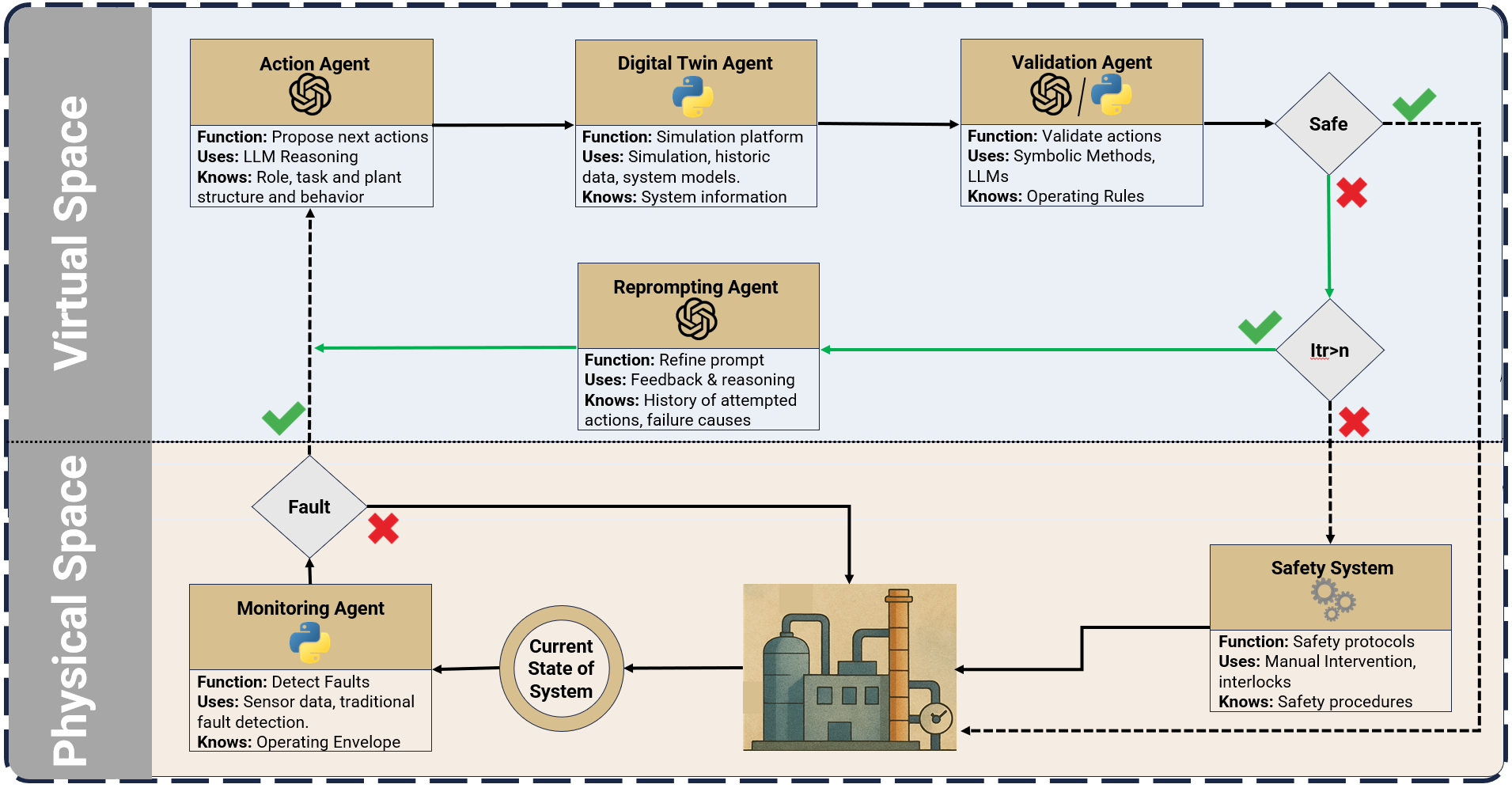}
    \caption{Five-component agentic pipeline: \textbf{\textit{monitoring}}, \textbf{\textit{action}}, \textbf{\textit{simulation}}, \textbf{\textit{validation}}, and \textbf{\textit{reprompting}}.}
    \label{fig:methodology_pipeline}
\end{figure}

\noindent
We conceptualize autonomous plant operation as a modular, five‐component pipeline that integrates continuous monitoring, symbolic state abstraction via FSMs, iterative planning, validation, and intelligent reprompting (Fig.~\ref{fig:methodology_pipeline}). Our architecture builds directly on established agentic control frameworks~\citep{ dycops2025}.

\subsection{Agent components}

\paragraph{\textbf{\textit{Monitoring agent}}}
Continuously ingests plant sensor streams and key performance indicators. \textbf{\textit{Monitoring agent}} maintains the current FSM state as an \emph{operating envelope} and checks in real time that measurements stay within prescribed bounds. When a deviation is detected, it initiates the planning loop by calling the \textbf{\textit{action agent}}.

\paragraph{\textbf{\textit{Action agent}}}
Receives the present FSM state, the desired operation state, and contextual metadata. \textbf{\textit{Action agent}} proposes a set of candidate control moves—discrete transitions or continuous set-point adjustments—intended to steer the system toward the goal while respecting envelope constraints.

\paragraph{\textbf{\textit{Simulation (digital twin)}}}
Each candidate move is executed off-line in a high-fidelity digital twin of the plant. \textbf{\textit{Simulation}} applies the proposed inputs to the process model and returns the ensuing trajectory, enabling risk-free assessment before real deployment.

\paragraph{\textbf{\textit{Validation agent}}}
Scrutinises the simulated outcome using rule-based checks and FSM semantics. If limits are violated, \textbf{\textit{validation agent}} issues a structured \texttt{rejection} report that pinpoints offending constraints; otherwise, it returns \texttt{valid}.

\paragraph{\textbf{\textit{Reprompting agent}}}
Invoked only after a rejection, \textbf{\textit{reprompting agent}} interprets the validator’s feedback and crafts a refined prompt for the \textbf{\textit{action agent}}, highlighting violations and suggesting corrective heuristics. The revised prompt triggers another planning iteration.

\subsection{Iterative control loop and safety override}

The cycle \textbf{\textit{action}}~$\rightarrow$~\textbf{\textit{simulation}}~$\rightarrow$~\textbf{\textit{validation}} (and, if necessary, \textbf{\textit{reprompting}}) repeats until a \texttt{valid} move is found or a predefined iteration budget is exhausted. If no satisfactory action emerges within this limit, a safety override hands control to a conservative fallback policy or a human operator, ensuring plant stability.

This modular, agentic design cleanly separates symbolic planning from continuous control, while the \textbf{\textit{reprompting agent}} provides feedback-driven adaptation beyond simple PID-style loops, forming the core of our autonomous fault-handling strategy.

\subsection{Agent‐orchestration frameworks}
\label{subsec:orchestration}

Large–language–model (LLM) \emph{agentic systems} typically require a
middleware layer that schedules dialogue turns, routes messages,
invokes external tools, and records shared state among otherwise
stateless model calls.  Frameworks such as \textsc{AutoGen}
\citep{wu2023autogen}, \textsc{LangChain}\,\citep{langchain}, and
\textsc{CrewAI}\,\citep{crewai2024} fulfill this role by exposing
\emph{agents} as first–class objects equipped with:

\begin{enumerate}
    \item \textbf{Interaction policies}: declarative rules that govern
          which agent speaks next, enabling complex multi‐party
          protocols without hard-coding conversation paths.
    \item \textbf{Tool adapters}: wrappers that let an LLM delegate
          tasks—e.\,g., database queries, simulations, or PLC
          commands—to deterministic code and then consume the result
          as fresh context.
    \item \textbf{Memory and state}: shared key–value stores or
          vector databases that persist long-horizon context while
          respecting token budgets.
    \item \textbf{Execution safeguards}: timeouts, retries, and
          concurrency controls that bound latency and prevent runaway
          token consumption—essential in safety-critical industrial
          loops.
\end{enumerate}

Because these frameworks abstract away low-level plumbing, researchers
can prototype new coordination strategies (e.\,g., voting, bidding,
hierarchies) by re-configuring YAML or JSON manifests rather than
rewriting code, accelerating reproducibility across laboratories and
use cases.

\subsection{Prompt‐based agent definition}
\label{subsec:prompt_design}

Each \textbf{\textit{agent}} is instantiated through a structured
prompt template that specifies its \textsc{role}, \textsc{goal}, and
\textsc{task}.  Templates are populated
at runtime with instance variables (e.\,g., sensor readings, constraint
limits) and dispatched to the underlying LLM with conservative
sampling parameters (temperature $=0$, \texttt{top\_p}$=0.1$) to
favour reproducibility over creativity.  A typical prompt includes:

\begin{itemize}
    \item \textbf{System directive} — high-level instructions that
          anchor the agent’s persona and delimit its authority.
    \item \textbf{Context block} — structured facts (JSON or tables)
          injected verbatim to minimise hallucination.
    \item \textbf{Output schema} — a JSON schema or few-shot example
          that enforces machine-parsable replies.
    \item \textbf{Constraint reminders} — succinct lists of
          physical or logical limits to suppress unsafe proposals.
\end{itemize}

When an action proposed by an \textbf{\textit{action agent}} is
flagged unacceptable by a downstream checker, the raw feedback is
appended to a new \textsc{system} message and re-issued to a dedicated
\textbf{\textit{reprompting agent}}.  This \emph{validator-aware
reprompting} allows the LLM to iteratively refine its answer without
human intervention, a pattern shown to reduce error rates in recent
multi-agent studies\,\citep{wu2023autogen}.  Token budgets are capped
at 512 per turn; lengthy constants are replaced by placeholder tags
that the orchestration layer expands on arrival, keeping latency
within real-time control bounds.

Together, the orchestration middleware and disciplined prompt
engineering provide a generic, reusable substrate on which
\textbf{\textit{monitoring}}, \textbf{\textit{action}},
\textbf{\textit{simulation}}, \textbf{\textit{validation}}, and
\textbf{\textit{reprompting}} agents interact safely and efficiently.

%
%
%
%
%
\section{Case studies}
\label{sec:case_studies}

In the following case studies, we implement two instantiations of the previously described agentic framework to evaluate its effectiveness across complementary dimensions of autonomy. The first realization focuses on symbolic planning using FSM traversal, testing the system's ability to generate valid state transitions under fault scenarios. The second realization targets continuous control in a dynamic process setting, assessing the framework's capacity for real-time decision-making and adaptive refinement of control policies through the reprompting loop. Together, these studies probe the framework's generality, robustness, and potential for deployment in safety-critical industrial environments.

\subsection{Case study 1: FSM traversal for state recovery planning}

Finite State Machines (FSMs) provide a structured abstraction for modeling discrete system modes and their allowable transitions. Their symbolic clarity and interpretability make them especially well-suited for recovery and reconfiguration tasks in industrial automation, where traceable behavior is essential.

In this case study, we investigate whether LLMs can autonomously generate valid transition sequences within a predefined FSM, starting from a initial state (representing faulty operation) and progressing to a designated operation state (representing normal operation) without the use of hardcoded planning algorithms. Unlike control-focused tasks that rely on dynamic models and numeric optimization, this setting emphasizes symbolic planning, multi-step reasoning, and instruction-following. The challenge lies not in executing actions, but in navigating abstract state graphs through valid, goal-directed sequences.

\begin{figure}[ht]
\centering
\fbox{\includegraphics[width=0.9\textwidth]{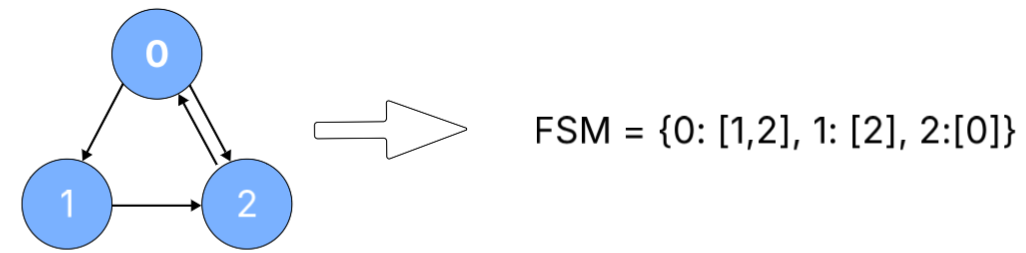}}
\caption{Illustration of a Finite State Machine encoded as a dictionary}
\label{fig:fsm_to_dict}
\end{figure}

To enable this capability, FSMs are encoded as Python dictionaries (e.g., \texttt{FSM = \{0: [1, 2], 1: [2], 2: [0]\}}) e.g. figure~\ref{fig:fsm_to_dict}, which we found to be more interpretable and reliably processed by LLMs than adjacency matrices or flat edge lists. This representation allows the agent to explicitly query permissible transitions and reason over symbolic structures in a way that aligns with the language-based interface of the LLM. The planning process is embedded within a validation-feedback loop, allowing the system to detect and correct invalid transitions through reprompting, without external supervision.

\subsubsection{Agentic framework for FSM traversal}

\begin{figure}[H]
\centering
\includegraphics[width=0.9\textwidth]{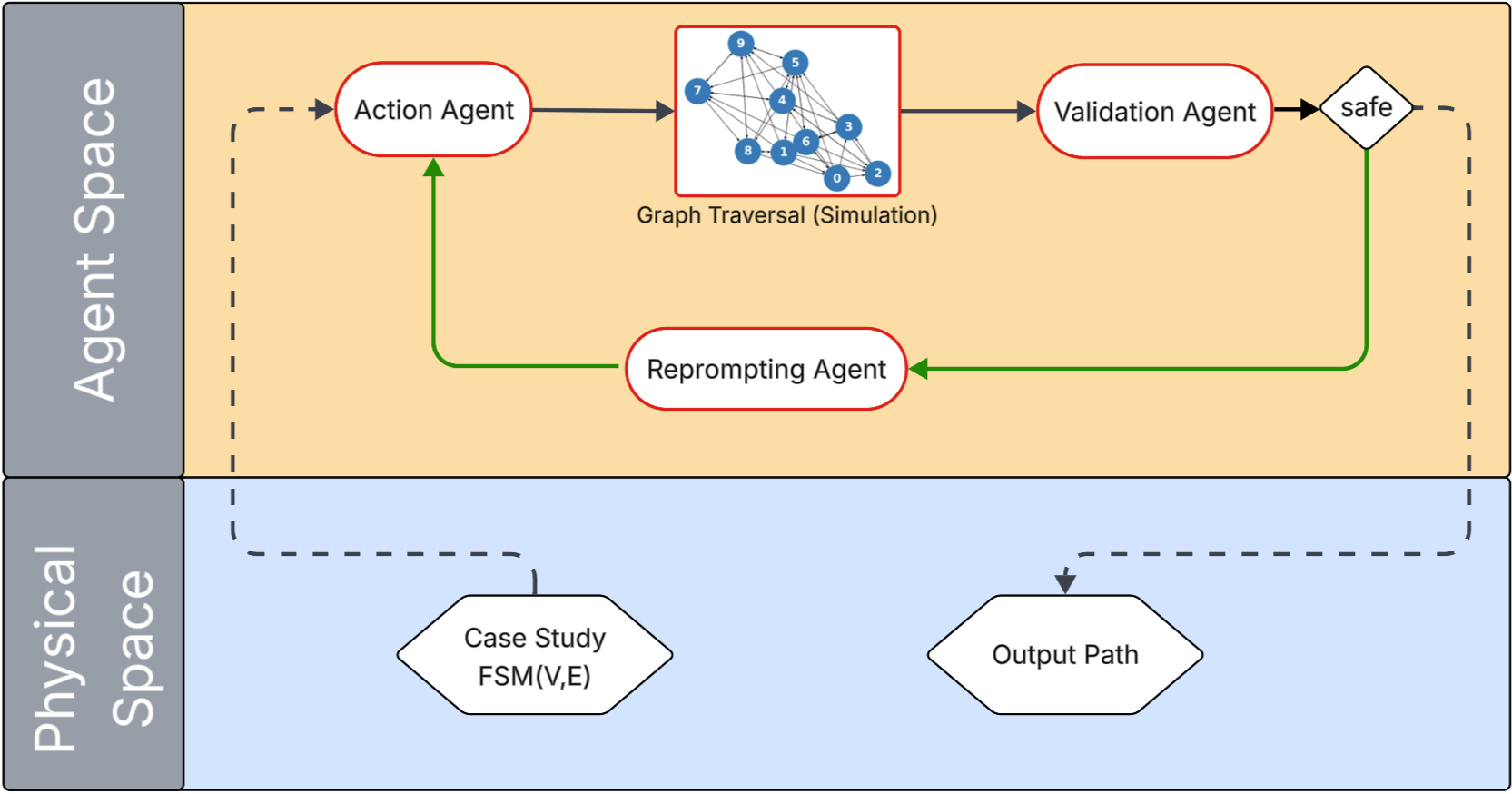}
\caption{Agent setup for FSM case study.}
\label{fig:fsm_sketch}
\end{figure}

To facilitate symbolic planning over FSMs, we deploy a modular multi-agent architecture that mirrors key cognitive functions—proposing, validating, and refining decisions—through discrete agent roles. This architecture is illustrated in Fig.~\ref{fig:fsm_sketch}.

\noindent The framework consists of the following components:

\begin{itemize}
    \item \textbf{Action Agent:} Given the initial and goal states, this agent proposes a candidate path composed of intermediate FSM states. The agent operates purely via natural language and symbolic reasoning, without access to ground-truth search algorithms or environment dynamics.

    \item \textbf{Simulation Agent:} Serves as a lightweight path executor by walking through the proposed state sequence. At each step, it checks whether the next state is reachable from the current one, using the FSM dictionary. It returns a structured execution trace along with any failure points encountered during traversal.

    \item \textbf{Validation Agent:} Interprets the output of the Simulation Agent to assess overall path feasibility. It identifies invalid transitions, diagnoses the point of failure, and routes the result back to the Reprompting Agent for targeted correction. 
    
    \item \textbf{Reprompting Agent:} In the event of validation failure, this agent modifies or augments the prompt fed to the Action Agent. The goal is to steer generation toward valid transitions without external labels or reinitializing the system.
\end{itemize}

\noindent This architecture supports a bounded number of refinement cycles (five in this case study), after which the last trajectory is selected. Notably, no simulation environment or plant dynamics are involved in this case study; the FSM itself constitutes the complete world model. While abstract, this formulation enables systematic evaluation of LLM capabilities in instruction following, symbolic planning, and self-correction. In future work, such FSMs may be derived from cyber-physical systems to bridge the gap between symbolic abstractions and real-world dynamics.

\subsubsection{Benchmark generation and evaluation metrics}
We begin by generating 20 random FSM instances for each (no. nodes, no. edges) pair in three difficulty category: \textit{easy}, \textit{medium}, and \textit{hard}. The \textit{easy} category includes FSMs with 4–6 nodes, the \textit{medium} category includes 10–15 nodes, and the \textit{hard} category includes 20–25 nodes. The number of edges for each FSM is selected to create denser graphs, typically using up to \( \binom{n}{2} \) edges. For smaller FSMs with 4 and 6 nodes, the edge count may include edges up to \( \binom{n}{3} \) to introduce higher complexity despite the smaller node set.

FSMs are constructed to ensure that no node is left disconnected. To enforce connectivity, we employ a graph generation algorithm that ensures each unconnected node is forcibly linked to at least one other node before proceeding with further random edge additions. This results in a globally connected FSM suitable for symbolic reasoning and path planning tasks. The algorithm used for generating such FSMs is described in Algorithm~\ref{alg:create_graph}.

\begin{algorithm}[ht]
\caption{Create FSM Graph with Global Connectivity}
\label{alg:create_graph}
\begin{algorithmic}[1]
\Require Number of nodes $N$, number of edges $R$
\Ensure Directed adjacency list $\text{adj}$ such that each node has at least one outgoing or incoming edge
\State Initialize adjacency list: $\text{adj} \gets [\,]$ for $N$ nodes
\State Set edge count $\text{edges} \gets 0$
\State Set unconnected node index $\text{non\_connected} \gets N-1$
\State Initialize flag $\text{flag} \gets \text{False}$
\While{$\text{edges} < R$}
    \State Randomly sample node $j \in \{0, \dots, N-1\}$
    \If{$\text{non\_connected} \geq 0$ and $\text{non\_connected} \neq j$}
        \State $i \gets \text{non\_connected}$ \Comment{Force connection for this node}
        \State $\text{flag} \gets \text{True}$
    \Else
        \State Randomly sample node $i \in \{0, \dots, N-1\}$
    \EndIf
    \If{$i \neq j$ and $j \notin \text{adj}[i]$}
        \State Add edge $(i \rightarrow j)$ to $\text{adj}$
        \State $\text{edges} \gets \text{edges} + 1$
        \If{$\text{flag} = \text{True}$}
            \State $\text{non\_connected} \gets \text{non\_connected} - 1$
            \State $\text{flag} \gets \text{False}$
        \EndIf
    \EndIf
\EndWhile
\end{algorithmic}
\end{algorithm}

Three LLM models are evaluated on the generated instances using up to five reprompting cycles per FSM. The following metrics are used to assess performance:
\begin{itemize}
    \item \textbf{First-pass accuracy:} Fraction of instances where the initially proposed path is valid.
    \item \textbf{Valid path accuracy:} Fraction of instances where a valid path is found within five reprompting attempts.
    \item \textbf{Average number of reprompts:} Mean number of reprompting cycles across all solved instances.
    \item \textbf{Path quality:} Average deviation from the shortest (optimal) path length.
\end{itemize}

\subsubsection{Results and observations}
\begin{figure}[h]
    \centering
    \begin{subfigure}[b]{0.3\textwidth}
        \fbox{\includegraphics[width=\textwidth]{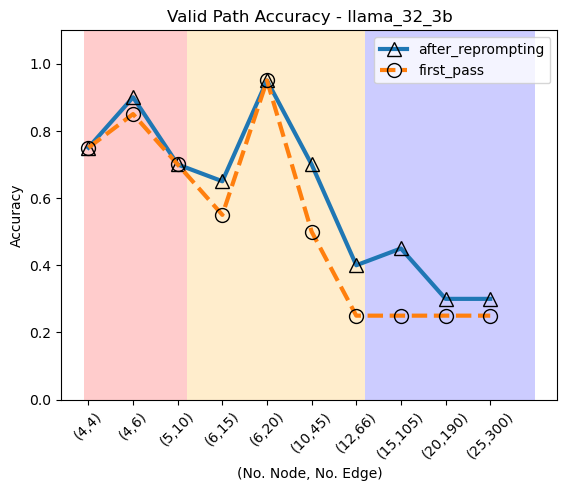}}
        \caption{Path Accuracy LLaMA - 3.2:3b}
        \label{fig:path_acc_llama}
    \end{subfigure}
    \hfill
    \begin{subfigure}[b]{0.3\textwidth}
        \fbox{\includegraphics[width=\textwidth]{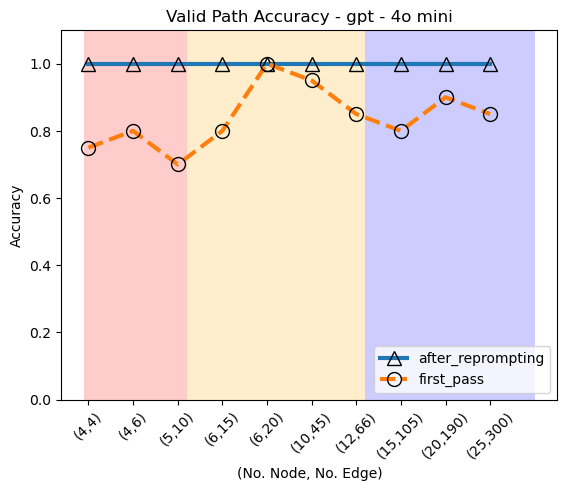}}
        \caption{Path Accuracy GPT-4o Mini}
        \label{fig:path_acc_4o_mini}
    \end{subfigure}
    \hfill
    \begin{subfigure}[b]{0.3\textwidth}
        \fbox{\includegraphics[width=\textwidth]{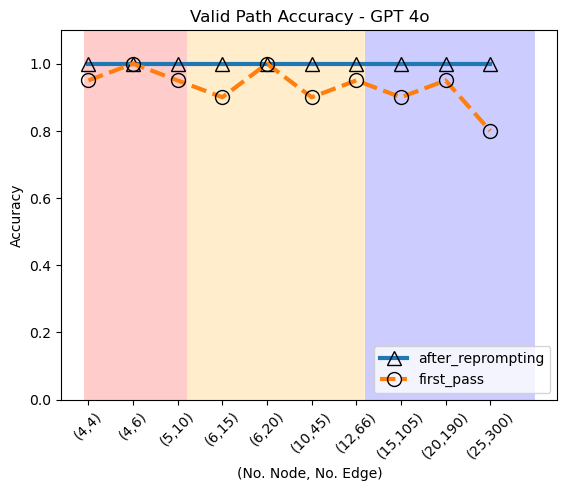}}
        \caption{Path Accuracy GPT-4o}
        \label{fig:path_acc_4o}
    \end{subfigure}
    \caption{Path accuracy comparison across models and FSM sizes.}
    \label{fig:Path_acc_fsm}
\end{figure}

The performance of the agentic planning framework was benchmarked across three models: LLaMA-3 (3.2B), GPT-4o-mini, and GPT-4o. As shown in Tables~\ref{table:llama_fsm_analysis},~\ref{table:gpt_4o_mini_fsm_analysis}, and~\ref{table:gpt_4o_fsm_analysis} and Figure~\ref{fig:Path_acc_fsm}, the OpenAI models substantially outperform the open-source LLaMA baseline in both planning accuracy and execution efficiency.

\begin{table}[!ht]
\centering
\caption{FSM Analysis - LLaMA}
\resizebox{\textwidth}{!}{
\begin{tabular}{|c|c|c|c|c|c|c|c|}
\hline
Nodes & Edges & First pass accuracy & Valid Path accuracy & Optimal Path Length & Deviation in Path Length & Avg. Reprompts & Avg. Time (s) \\\hline
4 & 4 & 0.85 & 0.90 & 1.85 & 0.55 & 0.55  & 6.011\\
4 & 6 & 0.75 & 0.75 & 2.20 & 0.60 & 1.25  &11.11\\
5 & 10 & 0.70 & 0.70 & 2.50 & 1.20 & 1.50 &12.70\\
6 & 15 & 0.55 & 0.65 & 2.65 & 1.60 & 1.95 &15.92\\
6 & 20 & 0.95 & 0.95 & 2.30 & 1.00 & 0.25 &3.97\\
10 & 45 & 0.50 & 0.70 & 3.50 & 2.35 & 1.70 &15.44\\
12 & 66 & 0.25 & 0.40 & 2.60 & 2.40 & 3.25 &27.19\\
15 & 105 & 0.25 & 0.45 & 2.60 & 2.25 & 3.05 & 27.76\\
20 & 190 & 0.25 & 0.30 & 2.70 & 2.35 & 3.55 &34.53\\
25 & 300 & 0.25 & 0.30 & 2.50 & 2.95 & 3.60 &36.86\\\hline
\end{tabular}}
\label{table:llama_fsm_analysis}
\end{table}

\begin{table}[ht]
\centering
\caption{FSM Analysis - GPT-4o Mini}
\resizebox{\textwidth}{!}{
\begin{tabular}{|c|c|c|c|c|c|c|c|}
\hline
Nodes & Edges & First pass accuracy & Valid Path accuracy & Optimal Path Length & Deviation in Path Length & Avg. Reprompts & Avg. Time (s)\\\hline
4 & 4 & 0.75 & 1.00 & 1.45 & 1.07 & 0.00   &1.66 \\
4 & 6 & 0.80 & 1.00 & 2.00 & 0.88 & 0.17   & 1.88\\
5 & 10 & 0.70 & 1.00 & 2.50 & 0.37 & 0.58  &3.11  \\
6 & 15 & 0.80 & 1.00 & 2.45 & 1.00 & 0.26  &2.13  \\
6 & 20 & 1.00 & 1.00 & 2.30 & 1.00 & 0.00  &1.66  \\
10 & 45 & 0.95 & 1.00 & 2.45 & 0.75 & 0.10 &2.17 \\
12 & 66 & 0.85 & 1.00 & 2.60 & 0.40 & 0.15 &1.95  \\
15 & 105 & 0.80 & 1.00 & 2.60 & 0.80 & 0.25 &2.53 \\
20 & 190 & 0.90 & 1.00 & 2.70 & 0.70 & 0.10 &2.30 \\
25 & 300 & 0.85 & 1.00 & 2.50 & 1.10 & 0.35 &2.96 \\\hline
\end{tabular}}
\label{table:gpt_4o_mini_fsm_analysis}
\end{table}

\begin{table}[h]
\centering
\caption{FSM Analysis - GPT-4o}
\resizebox{\textwidth}{!}{
\begin{tabular}{|c|c|c|c|c|c|c|c|}
\hline
Nodes & Edges & First pass accuracy & Valid Path accuracy & Optimal Path Length & Deviation in Path Length & Avg. Reprompts & Avg. Time (s)\\\hline
4 & 4 & 1.00 & 1.00 & 1.85 & 0.00 & 0.00 &3.29\\
4 & 6 & 0.95 & 1.00 & 2.20 & 0.05 & 0.05  &2.97\\
5 & 10 & 0.95 & 1.00 & 2.50 & 0.05 & 0.05 &3.26\\
6 & 15 & 0.90 & 1.00 & 2.65 & 0.20 & 0.10 &2.81 \\
6 & 20 & 1.00 & 1.00 & 2.30 & 0.40 & 0.00 &2.74\\
10 & 45 & 0.90 & 1.00 & 2.45 & 0.25 & 0.10  &2.88\\
12 & 66 & 0.95 & 1.00 & 2.60 & 0.35 & 0.10 &3.26\\
15 & 105 & 0.90 & 1.00 & 2.60 & 0.50 & 0.10 &3.25 \\
20 & 190 & 0.95 & 1.00 & 2.70 & 0.40 & 0.10 &3.52\\
25 & 300 & 0.80 & 1.00 & 2.50 & 0.45 & 0.25 &4.28\\\hline
\end{tabular}}
\label{table:gpt_4o_fsm_analysis}
\end{table}


GPT-4o achieves perfect \textit{valid path accuracy} across all FSM sizes, with first-pass success rates exceeding 90\% for most graph sizes. Its reprompt count remains low ($\leq$ 0.25 per instance), and the average deviation from the optimal path is under 0.5 steps, even on hard instances with over 300 edges. GPT-4o-mini exhibits similar robustness, with 100\% valid path accuracy across all tested graphs. While it requires slightly more reprompting in larger FSMs (up to 0.35 on average per instance), the generated paths remain near-optimal.

By contrast, LLaMA struggles with both instruction-following and multi-step symbolic planning. First-pass accuracy drops below 50\% for graphs with more than 10 nodes, and valid path recovery within five reprompts fails in many medium and hard instances. Average path deviations rise above 2 steps, and the reprompt count exceeds 3 per instance in the worst cases—indicating poor internal consistency during iterative reasoning.

Execution time trends further underscore this disparity. As shown in the final column of each table, GPT-4o and GPT-4o-mini complete planning cycles in under 4 seconds on average, including reprompt loops. In contrast, LLaMA inference times grow with graph complexity, exceeding 36 seconds for large FSMs. These delays reflect both slower local inference and greater reprompting effort.

A critical challenge encountered was that several models failed to follow instructions despite being given complete and unambiguous input. This behavior is not due to hallucination but instead points to weaknesses in instruction adherence. As the augmented prompts were instructed to avoid the previously explored paths, but most of the reprompting effort resulted in the previously explored path. Addressing such instruction-following deficiencies remains an important goal for improving LLM-based planners in symbolic and task-oriented settings.



\subsection{Case study 2: Continuous temperature control}

This case study investigates whether large language models (LLMs) can perform continuous control in a physical process environment. Specifically, we examine whether LLM-based agents can regulate temperature in a dual-heater setup under asymmetric disturbances. The control objective is to maintain the average temperature of two heaters at a desired setpoint, even when one heater is subjected to a constant cooling disturbance.

We use two Arduino-based Temperature Control Lab (TCLab) platform \citep{TCLab}, operating the heaters to simulate a larger space. A fan is positioned to continuously cool one heater, thereby emulating a fault condition (e.g., degraded actuator performance). This environment tests the ability of LLM agents to reason over continuous-valued control parameters and adaptively handle real-time disturbances.

\subsubsection{Experimental setup and multi-agent framework}

The setup consists of two physically separated TCLabs to minimize thermal interference. One heater is exposed to a fan-induced disturbance throughout the experiment, creating a persistent asymmetry in heater response.

\begin{figure}[h]
\centering
\includegraphics[width=0.9\textwidth]{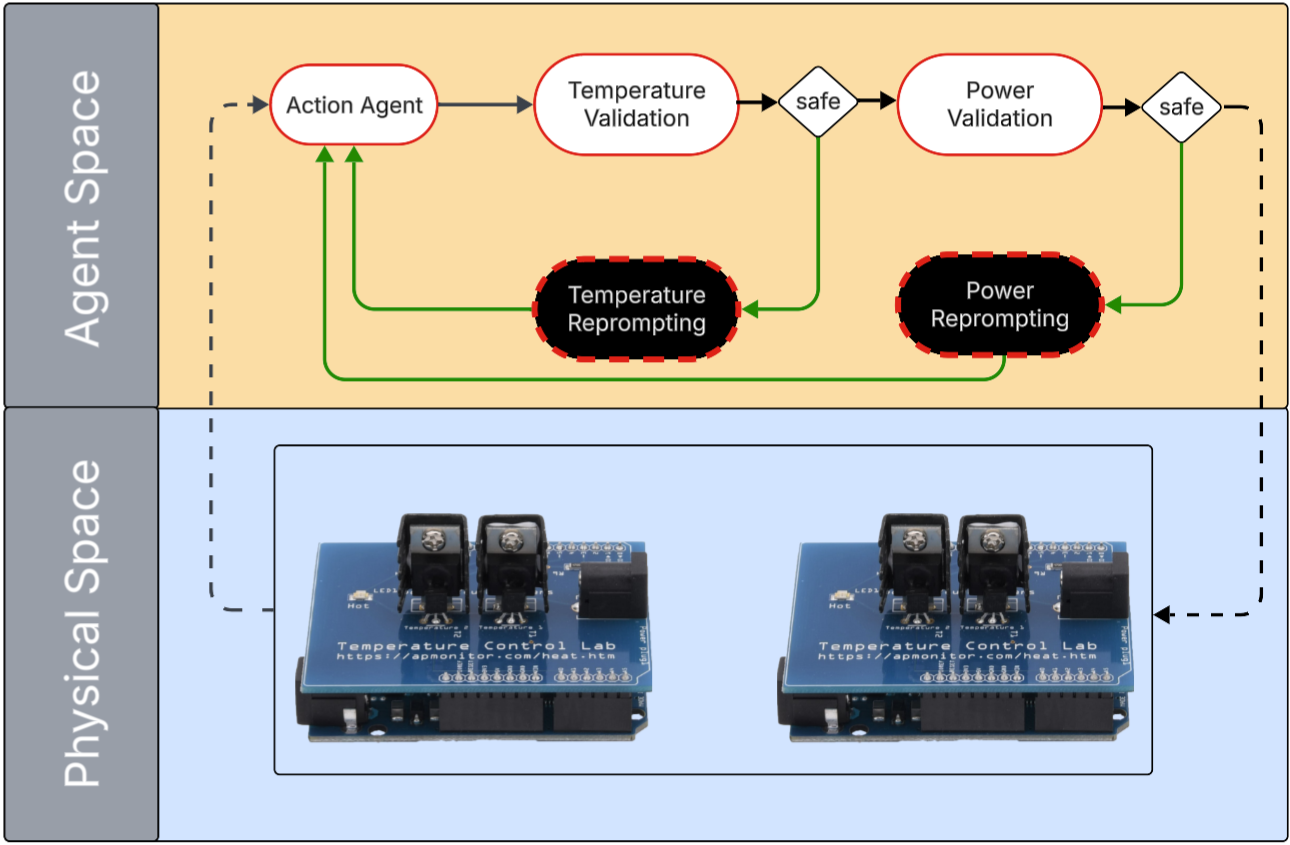}
\caption{Multi-agent framework for continuous temperature control.}
\label{fig:methodology_continuous}
\end{figure}

To manage this task, we adapt our multi-agent control framework as illustrated in Figure~\ref{fig:methodology_continuous}. A single \textit{Action Agent} proposes power values for both heaters. The proposed actions are validated and potentially revised through a two-stage reprompting mechanism:
\begin{itemize}
    \item The \textit{Temperature Validation Agent} assesses whether the control action brings the average temperature closer to the setpoint. If not, the \textit{Temperature Reprompting Agent} iteratively refines the actions.
    \item If the temperature criteria are satisfied, a \textit{Power Validation Agent} checks the physical plausibility of the commands (e.g., power within 0 to 0.3 W). Invalid values trigger the \textit{Power Reprompting Agent}.
\end{itemize}

Each inference cycle runs for a fixed number of iterations, after which the best available control action is applied to the plant. The entire loop is orchestrated using the CrewAI framework \cite{crewai2024}, with experiments conducted using both OpenAI and local LLaMA models.

\begin{figure}[h]
\centering
\fbox{\includegraphics[width=0.85\textwidth]{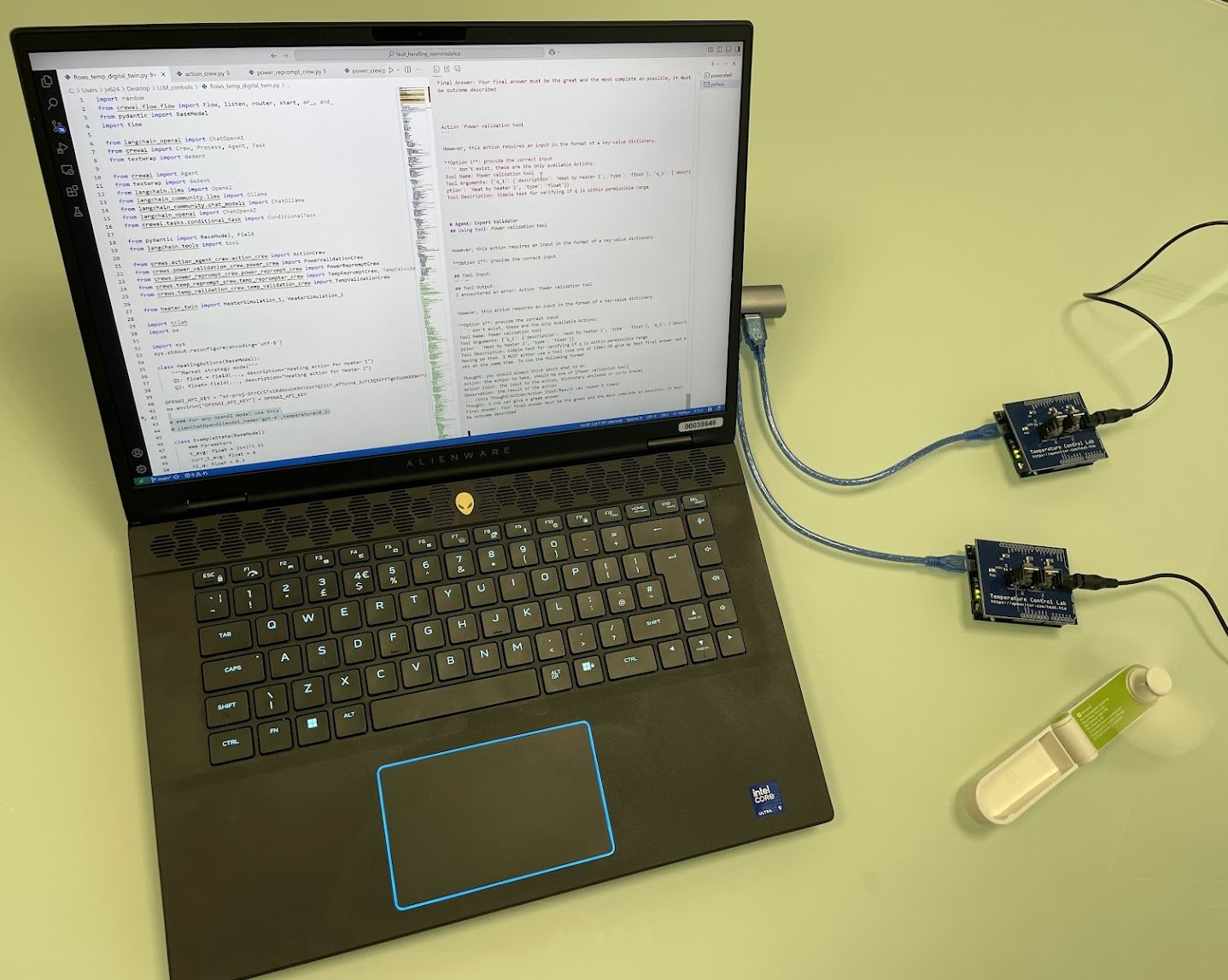}}
\caption{Physical experimental setup for dual-heater temperature control.}
\label{fig:setup_physical}
\end{figure}

Initial trials using the physical TCLab system (Figure~\ref{fig:setup_physical}) revealed several limitations, including variability in initial conditions, variable output latency, and low reproducibility due to ambient conditions. These challenges hampered rigorous evaluation.

To overcome this, we developed a digital twin of the two heaters in a Python environment, allowing deterministic simulation of heater dynamics under the same disturbance profiles. The emulated physcial system was paused using a fixed 30-second planning interval to accommodate LLM inference delays. This arrangement enabled repeatable, latency-aware experiments without compromising control fidelity.

We evaluate the performance of our framework across three deployment configurations, designed to assess the impact of inference environment and model access modality:

\begin{enumerate}
    \item \textbf{Physical plant with OpenAI model (cloud-based inference):} The control actions operate on real hardware (TCLab), with power commands generated via API calls to OpenAI-hosted models.
    
    \item \textbf{Digital twin with OpenAI model:} A simulated twin replicates the plant dynamics, while control decisions are still generated via cloud-based OpenAI models. This setup decouples physical variability from inference behavior.
    
    \item \textbf{Digital twin with LLaMA model (local inference):} The same digital twin is controlled using open-source LLaMA model running locally on an NVIDIA RTX 4060 GPU. This configuration evaluates the feasibility of on-premise, latency-sensitive LLM deployment.
\end{enumerate}

\begin{figure}[ht]
\centering
\begin{framed}
\begin{subfigure}[b]{0.8\textwidth}
\centering
\includegraphics[width=\textwidth]{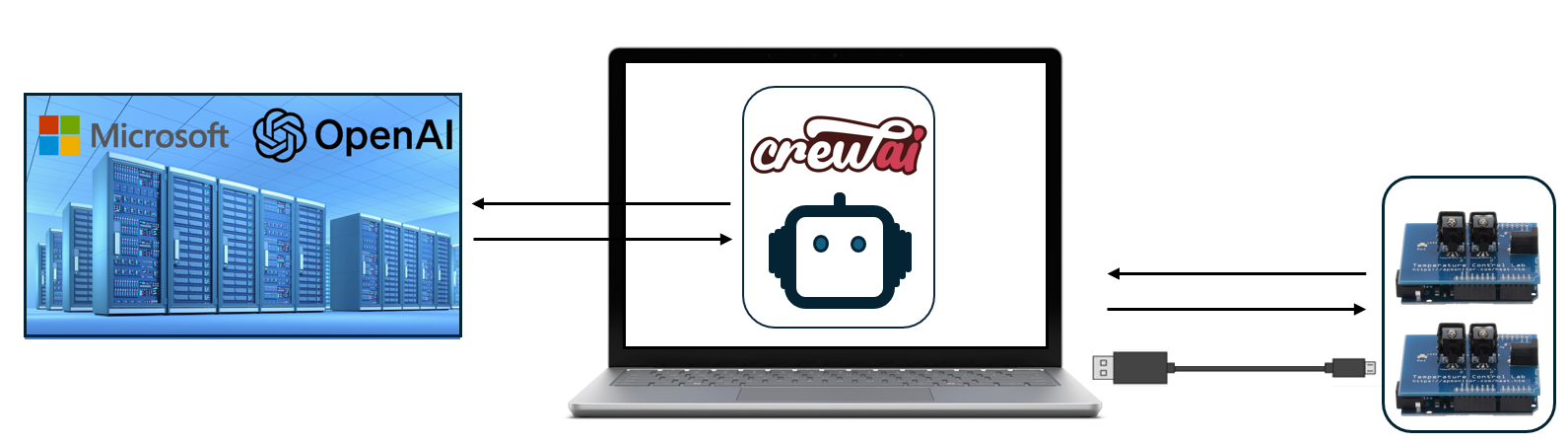}
\caption{Controlling the physical process via OpenAI API}
\label{fig:setup1}
\end{subfigure}

\vspace{0.1cm}

\begin{subfigure}[b]{0.8\textwidth}
\centering
\includegraphics[width=\textwidth]{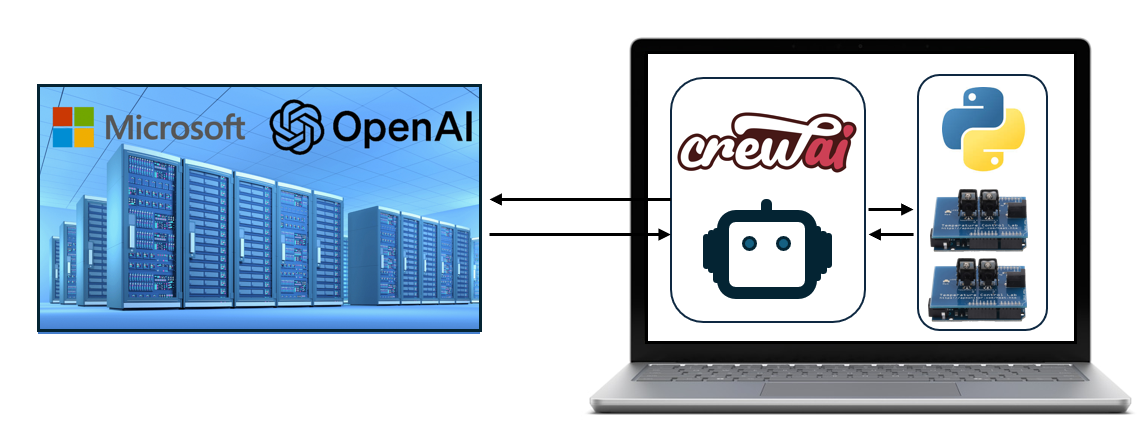}
\caption{Controlling the simulated plant via OpenAI API}
\label{fig:setup2}
\end{subfigure}

\vspace{0.1cm}

\begin{subfigure}[b]{0.6\textwidth}
\centering
\includegraphics[width=\textwidth]{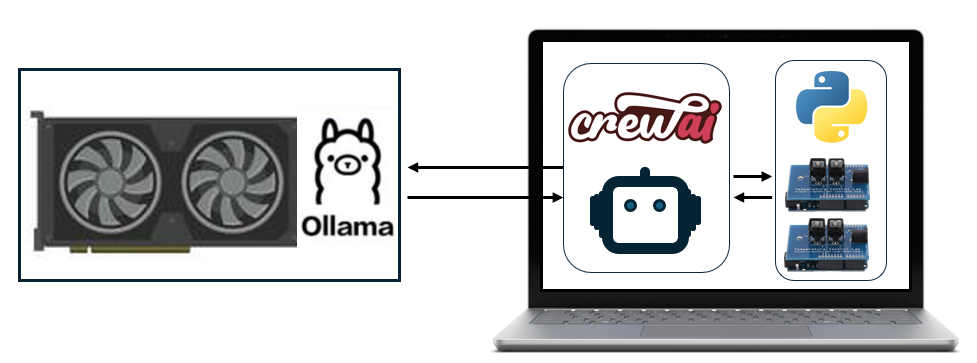}
\caption{Controlling the simulated plant with local LLaMA models}
\label{fig:setup3}
\end{subfigure}
\end{framed}
\caption{Three configurations used for performance evaluation.}
\label{fig:exp_setup_config}
\end{figure}

\subsubsection{Performance evaluation}

\textbf{Inference latency:}
Table~\ref{table:inference_time} summarizes the average inference times observed in both physical and simulated environments. The physical setup consistently exhibited higher and more variable latency compared to the digital twin. This discrepancy arises primarily due to conflicting information encountered during reprompting. Specifically, during the reprompting cycle, the LLM conditions its response on the current sensor readings (e.g., temperature and valve states). However, because the physical plant continues to evolve based on the previously issued control command (e.g., heaters remain active), the resulting state diverges from the expected one. This state mismatch introduces ambiguity into the LLM’s reasoning, often leading to an increased number of reprompts and consequently, longer inference durations.

\begin{table}[h] 
\centering
\caption{Inference time statistics}
\begin{tabular}{@{}|c|c|c|@{}}\hline 
Metric & Physical System & Digital Twin \\\hline 
Sample Count & 53 & 148 \\ 
Mean (s) & 34.18 & 12.49 \\ 
Std. Deviation (s) & 16.86 & 8.85 \\
Min/Max (s) & 10.58 / 72.69 & 5.12 / 93.00 \\ \hline 
\end{tabular}
\label{table:inference_time}
\end{table}

\textbf{Control accuracy:}
Control performance was evaluated using Time-Weighted Mean Absolute Error (TW-MAE) and Root Mean Square Error (RMSE), with lower values indicating more accurate regulation. As shown in Table~\ref{table:performance}, GPT-3.5 operating on the digital twin achieved performance closest to that of the PID baseline, demonstrating effective closed-loop control. GPT-4o exhibited a more refined understanding of system dynamics and produced smoother control actions, but required a larger number of reprompting cycles. The locally hosted LLaMA models showed inferior performance in both TW-MAE and RMSE, but remain attractive for offline deployment where low-latency inference and data privacy are prioritized.

\begin{table}[h]
\centering
\caption{Control performance metrics across different models}
\resizebox{\textwidth}{!}{
\begin{tabular}
{@{}|c|c|c|c|c|c|c|c|c|@{}}\hline
Model & PID Phy & GPT-3.5 Phy & PID DT & GPT-3.5 DT & GPT-4o-mini & GPT-4o & LLaMA-3.2 & LLaMA-3:8b\\\hline
TW-MAE & 0.7605 & 1.0277 & 0.2097 & 0.6182 & 1.2875 & 0.8245 & 1.8834 & 2.1232 \\
RMSE & 1.1382 & 1.2625 & 0.3349 & 1.0128 & 1.7401 & 1.2715 & 2.1704 & 2.4487 \\
Reprompts (Temp/Power) & - & 39/20 & - & 19/0 & 13/0 & 32/0 & 3/9 & 2/6 \\\hline
\end{tabular}
}
\label{table:performance}
\end{table}

\begin{figure}[H]
    \centering
    \begin{framed}
    \begin{subfigure}[b]{0.45\textwidth}
        \fbox{\includegraphics[width=\textwidth]{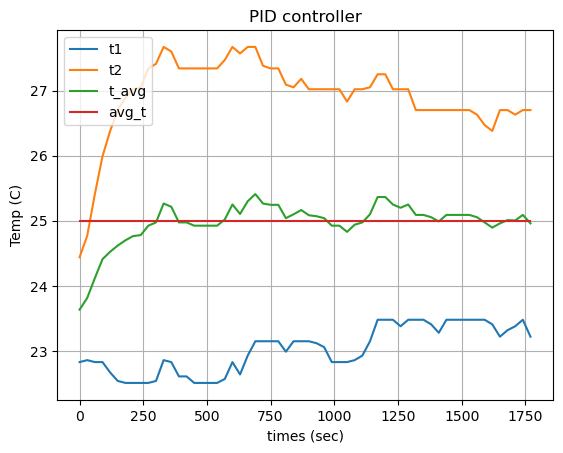}}
        \caption{PID Physical System}
        \label{fig:pid_physical_system}
    \end{subfigure}
    \hfill
    \begin{subfigure}[b]{0.45\textwidth}
        \fbox{\includegraphics[width=\textwidth]{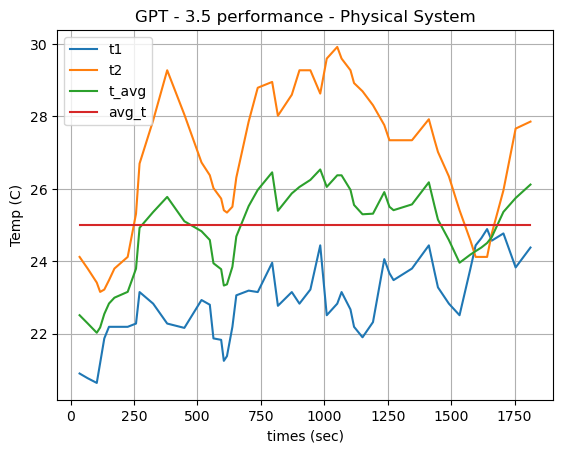}}
        \caption{GPT 3.5 - Physical System}
        \label{fig:gpt_35_physical}
    \end{subfigure}
    \end{framed}
    \caption{PID controller vs GPT 3.5 - Physical System}
    \label{fig:control_performance_physical}
\end{figure}

\begin{figure}[H]
    \centering
    \begin{framed}
    \begin{subfigure}[b]{0.32\textwidth}
        \centering
        \fbox{\includegraphics[width=\textwidth]{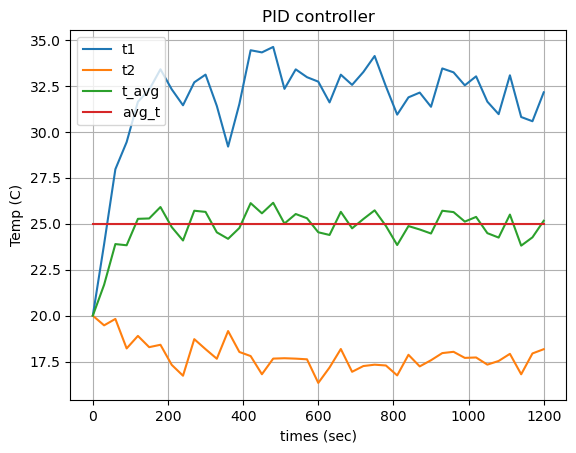}}
        \caption{PID controller}
        \label{fig:PID_controller}
    \end{subfigure}
    \hfill
    \begin{subfigure}[b]{0.32\textwidth}
        \centering
        \fbox{\includegraphics[width=\textwidth]{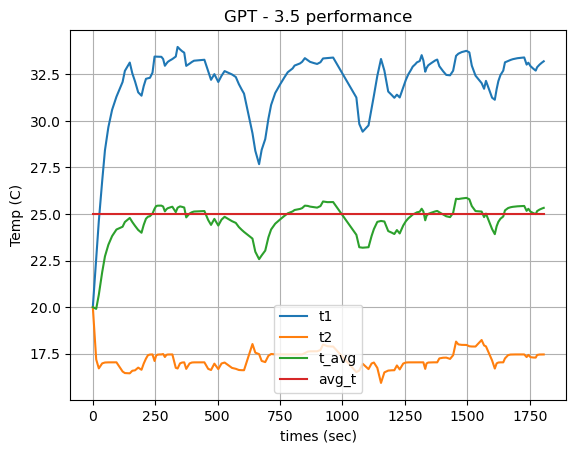}}
        \caption{GPT 3.5 performance}
        \label{fig:gpt35}
    \end{subfigure}
    \hfill
    \begin{subfigure}[b]{0.32\textwidth}
        \centering
        \fbox{\includegraphics[width=\textwidth]{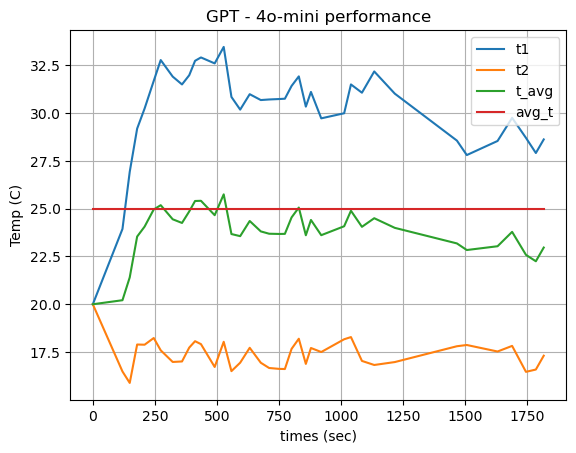}}
        \caption{GPT 4o mini performance}
        \label{fig:gpt4o_mini}
    \end{subfigure}

    \vskip\baselineskip 
    
    \begin{subfigure}[b]{0.32\textwidth}
        \centering
        \fbox{\includegraphics[width=\textwidth]{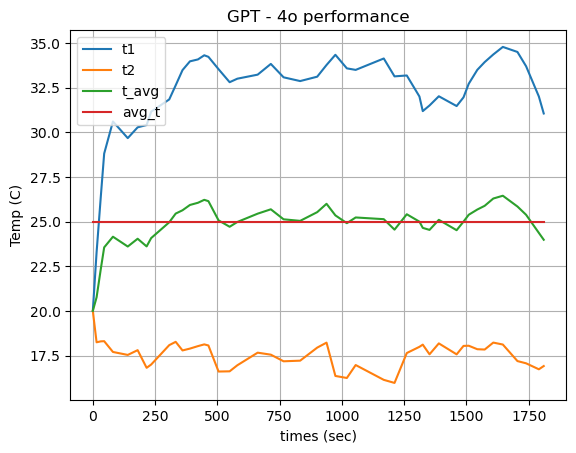}}
        \caption{GPT 4o performance}
        \label{fig:gpt4o}
    \end{subfigure}
    \hfill
    \begin{subfigure}[b]{0.32\textwidth}
        \centering
        \fbox{\includegraphics[width=\textwidth]{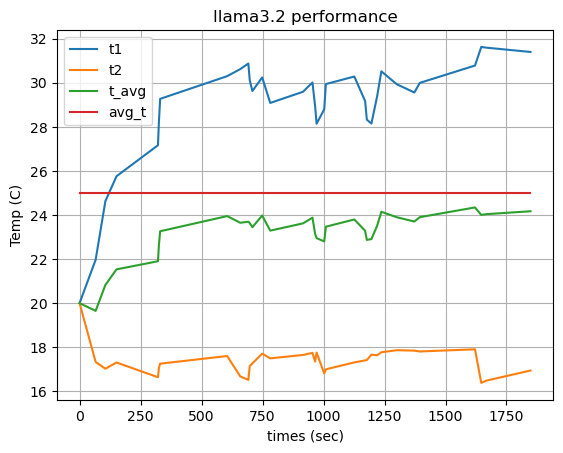}}
        \caption{LLAMA 3.2 performance}
        \label{fig:llama32}
    \end{subfigure}
    \hfill
    \begin{subfigure}[b]{0.32\textwidth}
        \centering
        \fbox{\includegraphics[width=\textwidth]{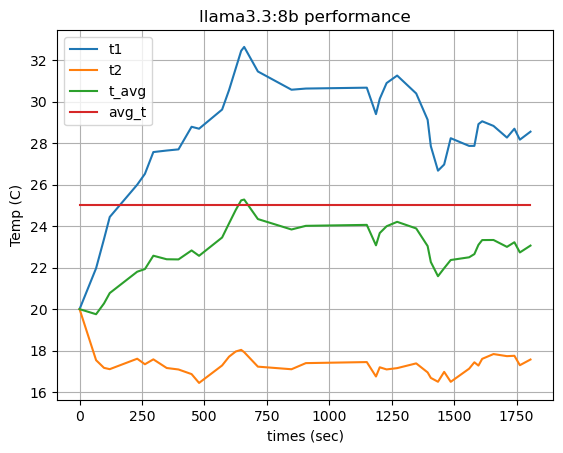}}
        \caption{LLAMA 3:8b performance}
        \label{fig:llama33_8b}
    \end{subfigure}
    \end{framed}
    \caption{Control performance of the system - Simulated in digital environment}
    \label{fig:control_performance}
\end{figure}

\textbf{Reasoning limitations:}
LLMs occasionally produced physically implausible approximations. For instance, Example~\ref{box:equal_power} demonstrates a failure mode in which the LLM naively distributes equal power to both heaters without accounting for their current temperature states, leading to overheating of the already hotter heater. In another case, shown in Example~\ref{box:forgot_heat_loss} illustrates a scenario in which the LLM neglects heat losses, resulting in overly optimistic temperature predictions. Additionally, Example~\ref{box:ode_solving}, the model attempts to solve an ODE using a single finite-difference step, instead of having smaller steps between the two states, thereby misrepresenting the system's dynamics. These behaviors indicate limitations in physical modeling fidelity. Table~\ref{table:odeint_vs_approx} quantifies how such inaccuracies accumulate with longer time horizons, emphasizing the importance of grounding LLM predictions in validated simulation feedback.

\begin{table}[h]
\centering
\caption{LLM-predicted vs. ground truth temperature over time.}
\resizebox{\textwidth}{!}{
\begin{tabular}{@{}|c|c|c|c|c|c|c|c|@{}}\hline
Time (s) & ODE (K) & LLM (K) & $\Delta$ & Time (s) & ODE (K) & LLM (K) & $\Delta$\\\hline
3 & 305.68 & 305.78 & 0.09 & 18 & 305.99 & 307.85 & 1.86\\
6 & 305.75 & 306.19 & 0.44 & 21 & 306.05 & 308.27 & 2.22\\
9 & 305.81 & 306.61 & 0.80 & 24 & 306.11 & 308.69 & 2.58\\
12 & 305.87 & 307.02 & 1.15 & 27 & 306.16 & 309.10 & 2.94\\
15 & 305.94 & 307.44 & 1.51 & 30 & 306.21 & 309.52 & 3.31\\\hline
\end{tabular}
}
\label{table:odeint_vs_approx}
\end{table}

\begin{examplebox}[label=box:equal_power]{LLM distributes power equally}
\begin{tabular}{l}

Total Power Input ($Q_{total}$) = $Heater_1\_Power$ + $Heater_2\_Power$\\

$Q_{total}$ = 0.25W + 0.25W = 0.5W\\

Next, we need to calculate the total heat energy input ($Q_{total}$) over the \\
30-second period:\\

$Q_{total}$ = Power * Time\\

$Q_{total}$ = 0.5W * 30s = 15J\\

Using the formula for heat energy (Q = m * $C_p$ * $\Delta$ T), we can calculate the \\

temperature change for each heater:\\

Q = m * $C_p$ * $\Delta$ T\\

$\Delta$ T = Q / (m * $C_p$)

\end{tabular}
\end{examplebox}

\begin{examplebox}[label=box:forgot_heat_loss]{LLM failing to account for heat losses}
\begin{tabular}{l}

The first step is to calculate the energy input for each heater. \\

Using the formula Q = Pt (where Q is the energy input, P is power, \\

and t is time), we get the energy input for heater 1 as 0.291W * 30s = 8.73J \\

and for heater 2 as 0.0624W * 30s = 1.872J.\\

Next, we can calculate the change in temperature for each heater using the \\

energy balance equation:\\

Q = m * $C_p$ * $\Delta$ T\\

Where Q is the energy input, m is the mass of the heater, $C_p$ is the heat \\

capacity, and $\Delta$ T is the change in temperature.
\end{tabular}
\end{examplebox}

\begin{examplebox}[label=box:ode_solving]{LLM incapable of solving ODEs}
\begin{tabular}{l}

- For Heater 1:\\

$Q_1$ = m * $C_p$ * ($T_1$ - $T_a$) \\

\hspace{1.75em}= 0.004 * 500 * (305.61482823373746 - 293.15) = 20.8590146189 J\\

$Q_1\_net$ = $Q_1$ - A * U * ($T_1$ - $T_a$) - $\epsilon$ * $\sigma$ * A * ($T_1^4$ - $T_a^4$) \\

\hspace{3.4em}= 20.8590146189 - 1.2$e^{-3}$ * 10 * (305.61482823373746 - 293.15) \\

\hspace{4.5em}- 0.9 * 5.67$e^{-8}$ * 1.2$e^{-3}$ * ($305.61482823373746^4$ - $293.15^4$) \\

\hspace{3.4em}= 20.8590146189 - 0.00421624556 - 0.166265144 = 20.6885332293 J\\

$T_1\_pred$ = (($Q_1\_net$ / (m * $C_p$)) + $T_a$) \\

\hspace{3.7em}= (20.6885332293 / (0.004 * 500)) + 293.15 = 312.255066458K\\

- For Heater 2:\\

$Q_2$ = m * $C_p$ * ($T_2$ - $T_a$) \\

\hspace{1.75em}= 0.004 * 500 * (289.4683954730248 - 293.15) = -14.5256170519 J\\

$Q_2\_net$ = $Q_2$ - A * U * ($T_2$ - $T_a$) - $\epsilon$ * $\sigma$ * A * ($T_2^4$ - $T_a^4$) \\

\hspace{3.4em}= -14.5256170519 - 1.2$e^{-3}$ * 10 * (289.4683954730248 - 293.15) \\

\hspace{4.5em}- 0.9 * 5.67$e^{-8}$ * 1.2$e^{-3}$ * ($289.4683954730248^4$ - $293.15^4$) \\

\hspace{3.4em}= -14.5256170519 + 0.00421624556 + 0.166265144 = -14.3551356623 J\\

$T_2\_pred$ = (($Q_2\_net$ / (m * $C_p$)) + $T_a$) \\

\hspace{3.7em}= (-14.3551356623 / (0.004 * 500)) + 293.15 = 284.721416245K

\end{tabular}
\end{examplebox}

\section{Discussion}
\label{sec:discussion}

While our study demonstrates promising results across both symbolic and continuous control domains and underscores the capabilities of our proposed agentic operation framework, several limitations must be acknowledged regarding the scope, realism, and generalisability of the findings. The physical experiments were conducted on lab-scale systems with limited process complexity, slow dynamics, and restricted operating envelopes. These settings, while suitable for controlled evaluations, do not capture the full spectrum of variability, disturbances, or coupled unit interactions present in industrial environments. In particular, continuous control was performed for tracking a single setpoint, and fault recovery scenarios were defined using abstract finite-state machines (FSMs) rather than derived from actual plant data, alarm sequences, or interlock events. As such, the system's ability to generalise to unstructured, high-dimensional, or multi-modal plant state representations remains untested.

Another important constraint lies in the system's reliance on commercial cloud-based LLM inference (e.g., via OpenAI APIs). Although this enables access to powerful general-purpose models, it introduces practical barriers to deployment in safety-critical or latency-sensitive industrial systems. Cloud inference adds non-deterministic latency, potential network failure modes, and recurring usage costs. Furthermore, the reliance on closed LLMs could be unsuitable for industrial processes such as chemical plants, where data sovereignty and auditability are non-negotiable. Current token context limits also restrict the amount of historical process data or memory that can be encoded in a single planning session, thereby constraining multi-turn reasoning or root-cause analysis across extended horizons.

Practical deployment of LLM-based agents will require deeper integration with existing industrial control hierarchies such as supervisory control and data acquisition (SCADA) systems and distributed control systems (DCS). Within such architectures, LLM agents may initially serve as non-critical supervisory advisors—e.g., recommending fault mitigation steps, flagging anomalous operating regimes, or assisting with setpoint transitions—before being entrusted with closed-loop authority. To ensure robustness and interpretability in these roles, future systems should incorporate domain-specific knowledge via lightweight fine-tuning, retrieval-augmented generation (RAG), or hybrid symbolic–neural memory models. Additionally, coupling LLM agents with physics-informed digital twins will enable grounded reasoning about process dynamics, constraint feasibility, and what-if counterfactuals.

From a systems perspective, future frameworks must move beyond loosely structured prompt chains to adopt standardised agent communication protocols that support modularity, introspection, and verifiability. Recent proposals such as the Model Context Protocol (MCP) offer a promising abstraction for formalising agent state, memory, tool interfaces, and dialogue control using serialisable schemas. This type of structured orchestration is essential for enabling composable, multi-agent architectures that are compatible with industrial software engineering practices. Finally, deployment in regulated sectors will require adherence to standards such as ISA-95, IEC 61511, and NIST SP 800-series for functional safety, cybersecurity, and traceability. Ensuring that LLM agents meet these requirements—e.g., through deterministic fallback policies, formal verification of output schemas, and system-level fault isolation—will be critical for their adoption in production environments.

\section{Conclusion}
\label{sec:conclusion}
This study presents a unified, agentic framework leveraging large language models for dual modes of industrial control: continuous actuation under physical constraints and symbolic planning over abstract finite state machines. By incorporating validation and reprompting agents, we enable fault-aware decision-making without explicit control laws or supervised training, using only task descriptions in natural language.

The system demonstrates strong performance in regulating temperature using two physically decoupled heaters, even under persistent asymmetric disturbances, and plans valid recovery paths in symbolic state spaces of increasing complexity. While classical PID control remains optimal for latency-constrained loops, our results suggest that LLMs can serve as robust, interpretable reasoning engines within hybrid architectures, particularly in supervisory or advisory roles.

By combining abstract planning with grounded actuation, this work lays the foundation for LLM-based autonomous operators capable of reasoning, correcting, and adapting to novel situations in complex industrial environments.
\section*{CRediT authorship contribution statement}

Javal Vyas: Writing – original draft, Writing – review \& editing, Methodology.

Mehmet Mercang\"{o}z: Conceptualization, Writing – review \& editing, Funding Acquisition.

\section*{Conflict of Interest}
The authors declare that they have no known competing financial
interests or personal relationships that could have appeared to influence the work reported in this paper

\section*{Data and Code Availability}
The source code is available at the following GitHub repository: \href{https://github.com/AISL-at-Imperial-College-London/multi\_agent\_llm\_control}{https://github.com/AISL-at-Imperial-College-London/multi\_agent\_llm\_control}

\section*{Declaration of generative AI and AI-assisted technologies in the writing process}

During the preparation of this work the author(s) used ChatGPT 4o in order to improve the readability of the paper. After using this tool/service, the author(s) reviewed and edited the content as needed and take(s) full responsibility for the content of the published article.

\section*{Acknowledgments}
Financial support from ABB for the Autonomous Industrial Systems Laboratory at Imperial College London is gratefully acknowledged.

\bibliography{cas-refs}
\pagebreak
\appendix
\section{Case study: FSM traversal}
\begin{tcolorbox}[title=Action Agent - Agent Prompt, colback=gray!20, colframe=orange!60!black, fontupper=\ttfamily\small, listing only, listing options={language=yaml, basicstyle=\ttfamily\footnotesize, breaklines=true}]
\begin{tabular}{l}
traversal\_expert:\\

  \hspace{1.5em}role: >\\
  
    \hspace{3em}Graph traversal expert\\
    
  \hspace{1.5em}goal: >\\
  
    \hspace{3em}Verify whether the final state is reachable and if return the best \\
    
    \hspace{3em}sequence of actions to achieve it.\\
    
  \hspace{1.5em}backstory: >\\
  
    \hspace{3em}The operator is a graph traversal expert who can determine the reachability \\
    
    \hspace{3em}of a state in a finite state machine and the sequence of states to \\
    
    \hspace{3em}reach it. \\

    \hspace{1.5em}skills: >\\

    \hspace{3em}- Graph traversal algorithms\\
    
    \hspace{3em}- Finite state machines\\
    
    \hspace{3em}- Breadth-first search\\
    
    \hspace{3em}- Depth-first search\\
    
    \hspace{3em}- Dijkstra's algorithm\\
    
    \hspace{3em}- A* search
\end{tabular}
\end{tcolorbox}

\begin{tcolorbox}[title=Action Agent - Task Prompt, colback=white, colframe=orange!60!black, fontupper=\ttfamily\small, listing only, listing options={language=yaml, basicstyle=\ttfamily\footnotesize, breaklines=true}]
\begin{tabular}{l}
traversal\_task:\\

  \hspace{1.5em}task: >\\
  
  \hspace{3em}Determine the reachability of a state in a finite state machine \\

  \hspace{3em}and the sequence of states to reach it.\\
  
  \hspace{1.5em}description:\\
  
    \hspace{3em}Can state {target\_state} be reached from {current\_state} given the adjacency \\
    
    \hspace{3em}list {graph}? Return a boolean value for this question \\
    
    \hspace{3em}{recommendation}\\

  \hspace{1.5em}agent: traversal\_expert\\

  \hspace{1.5em}expected\_output: >\\
  
  \hspace{3em}A boolean value indicating whether the target state can be reached from\\
  
  \hspace{3em}the current state given the adjacency list.\\

\end{tabular}
\end{tcolorbox}

\section{Case study: Continuous control prompt}

\begin{tcolorbox}[title=Action Agent - Agent Prompt, colback=gray!20, colframe=teal!60!black, fontupper=\ttfamily\small, listing only, listing options={language=yaml, basicstyle=\ttfamily\footnotesize, breaklines=true}]
\begin{tabular}{l}
operator:\\

  \hspace{1.5em}role: >\\
  
    \hspace{3em}Plant Operator\\
    
  \hspace{1.5em}goal: >\\
  
    \hspace{3em}Predict and adjust the heater outputs $q_1$ and $q_2$ to stabilize the system at \\
    
    \hspace{3em}an average temperature of {$t_{avg}$}C while minimizing the power used. \\
    
    \hspace{3em}The temperature is calculated as $(t_1 + t_2) / 2$ = {$t_{avg}$}C.\\ 
    
  \hspace{1.5em}backstory: >\\
    
    \hspace{3em}You are an experienced thermal plant operator with expertise in heat \\
    
    \hspace{3em}transfer and thermal systems. Your role is to ensure that the system \\
    
    \hspace{3em}remains stable and operates at the target temperature by predicting the \\
    
    \hspace{3em}necessary adjustments to the heater outputs. \\
    
    \hspace{3em}You have worked in the industry for several years and are adept at making \\
    
    \hspace{3em}decisions based on system behavior over time.
\end{tabular}
\end{tcolorbox}

\begin{tcolorbox}[title=Action Agent - Task Prompt, colback=white, colframe=teal!60!black, fontupper=\ttfamily\small, listing only, listing options={language=yaml, basicstyle=\ttfamily\footnotesize, breaklines=true}]
\begin{tabular}{l}
operator\_task:\\

  \hspace{1.5em}task: Determine the initial power settings for heater 1 and heater 2 to \\
  
  \hspace{5em}achieve an average system temperature of {$t_{avg}$} K.\\
  
  \hspace{1.5em}description: >\\
  
     \hspace{3em}Based on the current system state:\\
    
    \hspace{4em}- Current heater outputs: {$q_1$} W and {$q_2$} W\\
    
    \hspace{4em}- Current average temperature: {$curr\_t_{avg}$} K\\
    
    \hspace{4em}- Target average temperature: {$t_{avg}$} K\\
    
    \hspace{4em}- Heater temperature readings: {$t_1$} K, {$t_2$} K\\

    \hspace{3em}Heater performance is modeled using the following characteristics:\\
    
      \hspace{4em}- Heater output (Q) ranges from 0 to 1 W.\\
      
      \hspace{4em}- Heat capacity ($C_p$): 500 $J/kg-K$.\\
      
      \hspace{4em}- Surface area (A): $1.2 * 10^{-3} m^2$.\\
      
      \hspace{4em}- Mass (m): 0.004 kg.\\
      
      \hspace{4em}- Overall heat transfer coefficient (U): $10 W/m^2-K$.\\
      
      \hspace{4em}- Ambient temperature of the system is 293.15K.\\
      
      \hspace{4em}- Emissivity - 0.9\\
      
      \hspace{4em}- Stefan Boltzmann Constant - $5.67x10^{-8} W/m^2-K^{4}$\\
      
      \hspace{4em}- Ambient temperature - 293.15 K\\
      
      \hspace{4em}- The full energy balance includes convection and radiation terms.\\
      
      \hspace{4em}- The heater outputs must remain within the range of 0 W to 0.3 W.\\

    \hspace{3em}Your task is to propose initial values for $q_1$ and $q_2$ that will run for 30\\
    
    \hspace{4em}seconds while ensuring the following:\\
    
    \hspace{4em}- Average system temperature moves closer to {$t_{avg}$}K at the end of 30 \\
    
    \hspace{5em}seconds.\\
    
    \hspace{4em}- The power outputs remain within operational bounds: $0 W \leq q_1, q_2 \leq 0.3 W$.\\
    
    \hspace{4em}- Use available data to calculate the next state and propose values that \\
    
    \hspace{5em}balance efficiency and safety.\\
    
    \hspace{4em}- Output the temperatures after 30 seconds ($pred\_t_1, pred\_t_2$) along with\\
    
    \hspace{5em}the initial $q_1$ and $q_2$.\\ 

  \hspace{1.5em}agent: operator\\
  
  \hspace{1.5em}expected\_output: >\\
  
    \hspace{3em}A list containing the proposed values for heater power outputs: \\
    
    \hspace{3em}[$q_1$, $q_2$, $pred\_t_1$, $pred\_t_2$]
\end{tabular}
\end{tcolorbox}

\begin{tcolorbox}[title=Power Validation Agent - Agent Prompt, colback=gray!20, colframe=teal!60!black, fontupper=\ttfamily\small, listing only, listing options={language=yaml, basicstyle=\ttfamily\footnotesize, breaklines=true}]
\begin{tabular}{l}
power\_validator:\\

  \hspace{1.5em}role: >\\
  
    \hspace{3em}Expert Validator\\
    
  \hspace{1.5em}goal: >\\
  
    \hspace{3em}Verify if heaters output less than or equal to 30 and greater than or \\
    
    \hspace{3em}equal to 0.\\
    
  \hspace{1.5em}backstory: >\\
  
    \hspace{3em}Know how to use the validate the physical systems
\end{tabular}
\end{tcolorbox}

\begin{tcolorbox}[title=Power Validation Agent - Task Prompt, colback=white, colframe=teal!60!black, fontupper=\ttfamily\small, listing only, listing options={language=yaml, basicstyle=\ttfamily\footnotesize, breaklines=true}]
\begin{tabular}{l}
power\_validation\_task:\\

  \hspace{1.5em}task: >\\
  
    \hspace{3em}Validating agent outputs before passing it to the physical system and \\
    
    \hspace{3em}strictly output only one boolean value, True if validation tool return \\
    
    \hspace{3em}True else output False\\
  
  \hspace{1.5em}description: >\\
  
   \hspace{3em}Given the current heating outputs {$q_1$} and {$q_2$}, check if it is in the\\
   
   \hspace{3em}operating range of the system i.e. between {$lo_q$} and {$hi_q$}. \\
   
  \hspace{1.5em}note: >\\
  
    \hspace{3em}Strictly output only one boolean value, True if validation tool return\\
    
    \hspace{3em}True else output False in an array, as more outputs are costly for the \\
    
    \hspace{3em}LLMs .\\
    
    \hspace{1.5em}agent: power\_validator\\
    
    \hspace{1.5em}expected\_output: >\\
    
    \hspace{3em}Strictly output only one boolean value, \\
    
    \hspace{3em}True if validation tool returns True, else output False in an array
\end{tabular}
\end{tcolorbox}

\begin{tcolorbox}[title=Power Reprompter Agent - Agent Prompt, colback=gray!20, colframe=teal!60!black, fontupper=\ttfamily\small, listing only, listing options={language=yaml, basicstyle=\ttfamily\footnotesize, breaklines=true}]
\begin{tabular}{l}
power\_reprompter:\\

  \hspace{1.5em}role: Expert Fault Diagnosis Engineer\\
  
  \hspace{1.5em}goal: You have been assigned this task becuase power validation stage has \\
  
  \hspace{5em}failed and there is a power limit violation. \\
  
  \hspace{1.5em}backstory: You have worked in the industry as an expert plant operator and \\
  
  \hspace{7.5em}in depth knowledge of heat transfer.
\end{tabular}
\end{tcolorbox}

\begin{tcolorbox}[title=Power Reprompter Agent - Task Prompt, colback=white, colframe=teal!60!black, fontupper=\ttfamily\small, listing only, listing options={language=yaml, basicstyle=\ttfamily\footnotesize, breaklines=true}]
\begin{tabular}{l}
power\_reprompting\_task:\\

  \hspace{1.5em}task: > \\
  
    \hspace{3em}Keep the heater power outputs with the power limits.\\

  \hspace{1.5em}description: >\\
  
    \hspace{3em}Here the power limit has been violated and you need to make sure the \\
    
    \hspace{3em}power output is within the range of ({$lo_q$} W, {$hi_q$} W), while maintaining the \\
    
    \hspace{3em}average system temperature to be {$t_{avg}$}K. \\
    
    \hspace{3em}You are tasked with adjusting the power output of two heaters \\
    
    \hspace{3em}individually to stabilize the system at an average temperature of {$t_{avg}$} K. \\
    
    \hspace{3em}The system current temperatures are {$pred\_t_1$} K and {$pred\_t_2$} K
    Adjust the\\  
    
    \hspace{3em}power output of heaters at which they would run for 30 seconds and \\
    
    \hspace{3em}maintain the average temperature of {$t_{avg}$} K.\\
    
    \hspace{3em}Your goal is to adjust the heater power outputs to maintain an average \\
    
    \hspace{3em}temperature of {$t_{avg}$} K while adhering to the following constraints:\\
    
    \hspace{4em}- Current average temperature of the system is {$curr\_t_{avg}$} K.\\
    
    \hspace{4em}- Heater outputs must be between {$lo_q$} and {$hi_q$} watts.\\
    
    \hspace{4em}- Heater performance is modeled using the following characteristics:\\
    
      \hspace{4em}- Heater output (Q) ranges from 0 to 1 W.\\
      
      \hspace{4em}- Heat capacity ($C_p$): 500 $J/kg-K$.\\
      
      \hspace{4em}- Surface area (A): $1.2 * 10^{-3} m^2$.\\
      
      \hspace{4em}- Mass (m): 0.004 kg.\\
      
      \hspace{4em}- Overall heat transfer coefficient (U): 10 $W/m^2-K$.\\
      
      \hspace{4em}- Ambient temperature of the system is 293.15 K.\\
      
      \hspace{4em}- Emissivity - 0.9\\
      
      \hspace{4em}- Ambient temperatue - 293.15 K\\
      
      \hspace{4em}- Stefan Boltzmann Constant - $5.67x10^{-8} W/m^2-K^4$\\
      
      \hspace{4em}- The full energy balance includes convection and radiation terms.\\

    \hspace{3em}The next heater power outputs should make sure average temperature is\\
    
    \hspace{3em}moving towards {$t_{avg}$} K.\\
    
  \hspace{1.5em}Note: >\\
  
   \hspace{3em}Give me an array with float values for the next heater outputs, one value \\
   
   \hspace{3em}for $q_1$ and $q_2$ each. Anything other than that can cause system to fail.\\

  \hspace{1.5em}agent: power\_reprompter\\

  \hspace{1.5em}expected\_output: >\\
  
    \hspace{3em}Give me an array with float values for the next heater outputs [$q_1$,$q_2$]. \\
\end{tabular}
\end{tcolorbox}

\begin{tcolorbox}[title=Temperature Reprompter Agent 1 (Calculation Agent) - Agent Prompt, colback=gray!20, colframe=teal!60!black, fontupper=\ttfamily\small, listing only, listing options={language=yaml, basicstyle=\ttfamily\footnotesize, breaklines=true}]
\begin{tabular}{l}
temperature\_reprompter:\\

  \hspace{1.5em}role: >\\
  
    \hspace{3em}Intelligent Reprompter\\
    
  \hspace{1.5em}goal: >\\
  
    \hspace{3em}Calculate temperature of the system after 30 seconds.\\
    
  \hspace{1.5em}backstory: >\\
  
    \hspace{3em}A highly experienced troubleshooting expert in thermal systems. \\
    \hspace{3em}Skilled in proposing new strategies and adjustments when the primary \\
    
    \hspace{3em}optimization approach fails.\\
    
    \hspace{3em}You understand the physical dynamics of the system and can creatively \\
    \hspace{3em}search for solutions within the operational constraints.
\end{tabular}
\end{tcolorbox}

\begin{tcolorbox}[title=Temperature Reprompter Agent 2 (Heater 1 Agent) - Agent Prompt, colback=gray!20, colframe=teal!60!black, fontupper=\ttfamily\small, listing only, listing options={language=yaml, basicstyle=\ttfamily\footnotesize, breaklines=true}]
\begin{tabular}{l}
temperature\_reprompter\_1:\\

  \hspace{1.5em}role: >\\
  
    \hspace{3em}Intelligent Reprompter - Heater 1\\
    
  \hspace{1.5em}goal: >\\
  
    \hspace{3em}Adjust heater output to bring the average system temperature closer \\
    
    \hspace{3em}to {$t_{avg}$} C.\\
    
  \hspace{1.5em}backstory: >\\
  
    \hspace{3em}A highly experienced troubleshooting expert in thermal systems. Skilled \\
    
    \hspace{3em}in proposing adjustments when previous adjustments do not move towards \\
    
    \hspace{3em}the goal.\\
    
    \hspace{3em}You understand the physical dynamics of the system and can creatively \\
    
    \hspace{3em}search for solutions within the operational constraints.
\end{tabular}
\end{tcolorbox}

\begin{tcolorbox}[title=Temperature Reprompter Agent 3 (Heater 2 Agent) - Agent Prompt, colback=gray!20, colframe=teal!60!black, fontupper=\ttfamily\small, listing only, listing options={language=yaml, basicstyle=\ttfamily\footnotesize, breaklines=true}]
\begin{tabular}{l}
temperature\_reprompter\_2:\\

  \hspace{1.5em}role: >\\
  
    \hspace{3em}Intelligent Reprompter - Heater 1\\
    
  \hspace{1.5em}goal: >\\
  
    \hspace{3em}Adjust heater output to bring the average system temperature closer \\
    
    \hspace{3em}to {$t_{avg}$} C.\\
    
  \hspace{1.5em}backstory: >\\
  
    \hspace{3em}A highly experienced troubleshooting expert in thermal systems. Skilled \\
    
    \hspace{3em}in proposing adjustments when previous adjustments do not move towards \\
    
    \hspace{3em}the goal.\\
    
    \hspace{3em}You understand the physical dynamics of the system and can creatively \\
    
    \hspace{3em}search for solutions within the operational constraints.
\end{tabular}
\end{tcolorbox}
.
\begin{tcolorbox}[title=Temperature Reprompter Agent 1 (Calculation Agent) - Task Prompt, colback=white, colframe=teal!60!black, fontupper=\ttfamily\small, listing only, listing options={language=yaml, basicstyle=\ttfamily\footnotesize, breaklines=true}]
\begin{tabular}{l}
temp\_reprompting\_task\_1:\\

  \hspace{1.5em}task: Calculate temperature of heater 1 and heater 2 after 30 seconds.\\
  
  \hspace{1.5em}description: >\\
  
    \hspace{3em}Calculate the temperatures of heater 1 and heater 2 after 30 seconds with \\
    
    \hspace{3em}the {$q_1$} W and {$q_2$} W as power outputs and {$t_1$} K and {$t_2$} K as temperatures. \\
    
    \hspace{3em}From the current values of {$q_1$} W and {$q_2$} W gather information about how far\\
    
    \hspace{3em}the system is from average temperature of {$t_avg$} K.\\
    
    \hspace{3em}Below are the parameters describing the current state of the heaters and\\
    
    \hspace{3em}heater specification which should be used for energy. \\
    
    \hspace{3em}System Parameters:\\
    
    \hspace{4em}- Current temperatures: {$t_1$} K, {$t_2$} K\\
    
    \hspace{4em}- Current average temperature: {$curr\_t_{avg}$} K\\
    
    \hspace{4em}- Current heater outputs: {$q_1$} W, {$q_2$} W\\
    
    \hspace{4em}- Target average temperature: {$t_{avg}$} K\\
    
    \hspace{4em}- Heater output range: 0 to 0.3 watts\\

    \hspace{3em}Heater Specifications:\\
    
    \hspace{4em}- Heater output (Q) - 0 to 1 W\\
    
    \hspace{4em}- Heat capacity (Cp) - 500 $J/kg-K$\\
    
    \hspace{4em}- Surface area (A) - 1.2 * $10^{-3} m^2$\\
    
    \hspace{4em}- Mass (m) - 0.004 kg\\
    
    \hspace{4em}- Overall Heat Transfer Coefficient (U) - 10 $W/m^2-K$\\
    
    \hspace{4em}- Ambient temperature of the system is 293.15 K.\\
    
    \hspace{4em}- Emissivity - 0.9\\
    
    \hspace{4em}- Stefan Boltzmann Constant - $5.67x10^{-8} W/m^2-K^4$\\
    
    \hspace{4em}- Ambient temperatue - 293.15 K\\
    
    \hspace{4em}- The full energy balance includes convection and radiation terms.\\

    \hspace{3em}Task Details:\\
   \hspace{4em}- Assume the heaters will operate continuously for 30 seconds at the \\
   
   \hspace{5em}current outputs ({$q_1$}, {$q_2$}).\\
    
    \hspace{4em}- Use the heater specifications and energy balance equations to predict \\
    
    \hspace{5em}the resulting temperatures for heater 1 ($pred\_t_1$ K) and heater 2 \\
    
    \hspace{5em}($pred\_t_2$ K) after 30 seconds.\\

    \hspace{3em}The aim of this task is to find the heater output temperatures.
    Do not \\
    
    \hspace{3em}forget to consider to refer to a similar problem that you have solved \\
    
    \hspace{3em}previously. \\
  
  \hspace{1.5em}agent: temperature\_reprompter\\
  
  \hspace{1.5em}expected\_output: >\\
  
    \hspace{3em}An array with next heater temperature are: [$pred\_t_1$,$pred\_t_2$]
\end{tabular}
\end{tcolorbox}

\begin{tcolorbox}[title=Temperature Reprompter Agent 2 (Heater 1 Agent) - Task Prompt, colback=white, colframe=teal!60!black, fontupper=\ttfamily\small, listing only, listing options={language=yaml, basicstyle=\ttfamily\footnotesize, breaklines=true}]
\begin{tabular}{l}
temp\_reprompting\_task\_2:\\

  \hspace{1.5em}task: >\\
  
    \hspace{3em}Adjust the power of heater 1 {$q_1$} W based on the {$pred\_t_1$} K and {$pred\_t_2$} K \\
    
    \hspace{3em}values calculated in temp\_reprompting\_task\_1.\\
    
  \hspace{1.5em}description: >\\
  
    \hspace{3em}Determine the heater 1 power output which it should run for in the next \\
    
    \hspace{3em}state for 30 seconds after {$pred\_t_1$} K and {$pred\_t_2$} K \\
    
    \hspace{3em}are reached, taking into account the following conditions:\\
    
    \hspace{4em}- Average system temperature should move towards {$t_{avg}$} K.\\
    
    \hspace{4em}- Heater power output range: 0 to 0.3 watts.\\
    
    \hspace{4em}- The score with the current heater outputs {$q_1$} W and {$q_2$} W is {$avg\_score$}.\\
    
    \hspace{4em}- There is no cooling available for heaters, so the minimum power \\
    
    \hspace{5em}output can be 0 W.\\
    
    \hspace{4em}- The heating power output is between 0W and 0.3 W.\\

    \hspace{3em}- Heater performance is modeled using the following characteristics:\\
    
      \hspace{4em}- Heater output (Q) ranges from 0 to 1 W.\\
      
      \hspace{4em}- Heat capacity ($C_p$): 500 $J/kg-K$.\\
      
      \hspace{4em}- Surface area (A): $1.2 * 10^{-3} m^2$.\\
      
      \hspace{4em}- Mass (m): 0.004 kg.\\
      
      \hspace{4em}- Overall heat transfer coefficient (U): 10 $W/m^2-K$.\\
      
      \hspace{4em}- Ambient temperature of the system is 293.15 K.\\
      
      \hspace{4em}- Emissivity - 0.9\\
      
      \hspace{4em}- Stefan Boltzmann Constant - $5.67x10^{-8} W/m^2-K^4$\\
      
      \hspace{4em}- Ambient temperature - 293.15 K\\
      
      \hspace{4em}- The full energy balance includes convection and radiation terms.\\
      
      \hspace{4em}- The heater outputs must remain within the range of 0 W to 0.3 W.\\

    \hspace{3em}The aim of this task is to find the heater output for heater 1.\\
    
    \hspace{3em}Do not forget to consider to refer to a similar problem that you have \\
    
    \hspace{3em}solved previously.\\
    
  \hspace{1.5em}agent: temperature\_reprompter\_1\\
  
  \hspace{1.5em}note: >\\
  
    \hspace{3em}The heater outputs must remain within the range of 0 to 0.3 watts.\\
    
    \hspace{3em}Always ensure the system remains stable and provide only boolean response. \\
    
  \hspace{1.5em}expected\_output: >\\
  
    \hspace{3em}An array with the next heater 1 output: [$q_1$]
\end{tabular}
\end{tcolorbox}

\begin{tcolorbox}[title=Temperature Reprompter Agent 3 (Heater 2 Agent) - Task Prompt, colback=white, colframe=teal!60!black, fontupper=\ttfamily\small, listing only, listing options={language=yaml, basicstyle=\ttfamily\footnotesize, breaklines=true}]
\begin{tabular}{l}
temp\_reprompting\_task\_2:\\

  \hspace{1.5em}task: >\\
  
    \hspace{3em}Adjust the power of heater 1 {$q_1$} W based on the {$pred\_t_1$} K and {$pred\_t_2$} K \\
    
    \hspace{3em}values calculated in temp\_reprompting\_task\_1.\\
    
  \hspace{1.5em}description: >\\
  
    \hspace{3em} Your task is to determine the appropriate power output for heater 2 \\
    
    \hspace{3em}($q_2$) W which the heater 2 would run for the next 30 seconds while \\
    
    \hspace{3em}ensuring the following conditions:\\
    
    \hspace{4em}- The average system temperature should move closer to the target \\
    
    \hspace{5em}average temperature of {$t_{avg}$} K.\\
    
    \hspace{4em}- Heater output must remain within the range of 0 to 0.3 watts.\\
    
    \hspace{4em}- Use the temperature\_reprompter\_1 output $q_1$ as a fixed value for \\
    
    \hspace{4em}heater 1 ($q_1$ provided by temp\_reprompting\_task\_2).\\
    
    \hspace{4em}- There is no cooling available for heaters, so the minimum power \\
    
    \hspace{4em}output can be 0 W.\\
    
    \hspace{4em}- The heating power output is between 0 W and 0.3 W.\\

    \hspace{3em}- Heater performance is modeled using the following characteristics:\\
    
      \hspace{4em}- Heater output (Q) ranges from 0 to 1 W.\\
      
      \hspace{4em}- Heat capacity ($C_p$): 500 $J/kg-K$.\\
      
      \hspace{4em}- Surface area (A): $1.2 * 10^{-3} m^2$.\\
      
      \hspace{4em}- Mass (m): 0.004 kg.\\
      
      \hspace{4em}- Overall heat transfer coefficient (U): 10 $W/m^2-K$.\\
      
      \hspace{4em}- Ambient temperature of the system is 293.15 K.\\
      
      \hspace{4em}- Emissivity - 0.9\\
      
      \hspace{4em}- Stefan Boltzmann Constant - $5.67x10^{-8} W/m^2-K^4$\\
      
      \hspace{4em}- Ambient temperature - 293.15 K\\
      
      \hspace{4em}- The full energy balance includes convection and radiation terms.\\
      
      \hspace{4em}- The heater outputs must remain within the range of 0 W to 0.3 W.\\

    \hspace{3em}The aim of this task is to find the heater output for heater 1.\\
    
    \hspace{3em}Do not forget to consider to refer to a similar problem that you have \\
    
    \hspace{3em}solved previously.\\
    
    \hspace{1.5em}agent: temperature\_reprompter\_2\\
    
    \hspace{1.5em}note: >\\
    
    \hspace{3em}The heater outputs must remain within the range of 0 to 0.3 watts.\\
    \hspace{3em}Always ensure the system remains stable and provide heating output \\
    
    \hspace{3em}between 0 W and 0.3 W.\\
    
    \hspace{1.5em}expected\_output: >\\
    
    \hspace{3em}An array with the next heater 2 output and new average score: \\
    \hspace{3em}[$q_1$, $q_2$, $curr\_avg\_score$]
\end{tabular}
\end{tcolorbox}

\begin{tcolorbox}[title=Temperature Validation Agent - Agent Prompt, colback=gray!20, colframe=teal!60!black, fontupper=\ttfamily\small, listing only, listing options={language=yaml, basicstyle=\ttfamily\footnotesize, breaklines=true}]
\begin{tabular}{l}
average\_validator:\\

  \hspace{1.5em}role: >\\
  
    \hspace{3em}Thermal systems expert\\
    
  \hspace{1.5em}goal: >\\
  
    \hspace{3em}Only accept new heater outputs if they result in a closer average \\
    
    \hspace{3em}temperature compared to the previous one.\\
    
  \hspace{1.5em}backstory: >\\
  
    \hspace{3em}You are a control system specialist trained in optimization techniques, \\
    
    \hspace{3em}particularly gradient-based approaches. Your expertise lies in ensuring \\
    
    \hspace{3em}stability and efficiency by iteratively improving system performance. \\
    
    \hspace{3em}You assess the current and proposed heater outputs and ensure that only \\
    
    \hspace{3em}adjustments leading to an improvement are accepted.
\end{tabular}
\end{tcolorbox}

\begin{tcolorbox}[title=Temperature Validation Agent - Task Prompt, colback=white, colframe=teal!60!black, fontupper=\ttfamily\small, listing only, listing options={language=yaml, basicstyle=\ttfamily\footnotesize, breaklines=true}]
\begin{tabular}{l}
temp\_validation:\\

  \hspace{1.5em}task: Validate heater outputs.\\
  
  \hspace{1.5em}description: >\\
  
    \hspace{3em}Evaluate the power outputs of the two heaters to achieve an average \\
    
    \hspace{3em}system temperature closer to the target of {$t_{avg}$} K.\\
    
    \hspace{3em}Accept new heater outputs only if they reduce the deviation from the\\
    
    \hspace{3em}target average temperature. \\
    
    \hspace{3em}The current system parameters are:\\
    
    \hspace{4em}- Current temperatures: {$t_1$} K, {$t_2$} K\\
    
    \hspace{4em}- Current heater outputs: {$q_1$} W, {$q_2$} W\\
    
    \hspace{4em}- Target average temperature: {$t_{avg}$} K\\
    
    \hspace{4em}- Heater output range: 0 to 0.3 watts\\

    \hspace{3em}Heater Specifications:\\
    
    \hspace{4em}- Heater output (Q) - 0 to 1 W\\
    
    \hspace{4em}- Heater factor - 0.01 W/(\% heater)\\
    
    \hspace{4em}- Heat capacity (Cp) - 500 $J/kg-K$\\
    
    \hspace{4em}- Surface area (A) - $1.2 * 10^{-3} m^2$\\
    
    \hspace{4em}- Mass (m) - 0.004 kg\\
    
    \hspace{4em}- Overall Heat Transfer Coefficient (U) - 10 $W/m^{2}-K$\\

    \hspace{4em}Task Details:\\
    
    \hspace{4em}- Assume the heaters will operate continuously for 30 seconds at the\\
    
    \hspace{5em}current outputs ({$q_1$} W, {$q_2$} W).\\
    
    \hspace{4em}- Use the heater specifications and energy balance equations to predict \\
    
    \hspace{5em}the resulting temperatures for heater 1 ($pred\_t_1$ K) and heater 2 \\
    
    \hspace{5em}($pred\_t_2$ K) after 30 seconds.\\
    
    \hspace{4em}- Calculate the average temperature with $pred\_t_1$ K and $pred\_t_2$ K,\\
    
    \hspace{5em}$\frac{(pred\_t_1+pred\_t_2)}{2}$ K and compare it with the previous average score \\
    
    \hspace{5em}{avg\_score}.\\
  
  \hspace{1.5em}method:\\
  
    \hspace{4em}1. Calculate the average temperature of the system based on the \\
    
    \hspace{6em}predicted temperatures from current heater outputs ({$q_1$} W, {$q_2$} W).\\
    
    \hspace{4em}2. Compare the resulting average temperature with the target ({$t_{avg}$} K) \\
    
    \hspace{6em}and the previous average temperature.\\
    
    \hspace{4em}3. Allow the new heater outputs only if the new average temperature is \\
    
    \hspace{6em}closer to the target than the previous average temperature.\\
    
    \hspace{4em}4. Provide feedback indicating whether the adjustment was accepted \\
    
    \hspace{6em}or rejected.\\
    
    \hspace{4em}5. If the heating outputs are 0 W and you cannot improve on the score \\
    
    \hspace{6em}then accept 0 W and move on. \\
  
  \hspace{1.5em}note: >\\
  
    \hspace{3em}The heater outputs must remain within the range of 0 to 0.3 watts. \\
    
    \hspace{3em}Always ensure the system remains stable and provide only boolean \\
    
    \hspace{3em}response. \\
    
    \hspace{3em}Provide the output as True if the new heater outputs are accepted and \\
    
    \hspace{3em}False if the new heater outputs are rejected.\\
    
  \hspace{1.5em}agent: average\_validator\\

  \hspace{1.5em}expected\_output: Strictly output only one boolean value, True if validation \\
  
  \hspace{10.5em}tool returns True, else output False\\
  
\end{tabular}
\end{tcolorbox}

\end{document}